\documentclass[a4paper,fleqn]{cas-dc}

\usepackage[numbers]{natbib}
\usepackage{amsmath}
\usepackage{algorithmic}
\usepackage{graphicx}
\usepackage{textcomp}
\usepackage{xcolor}
\usepackage{bm}
\usepackage{amstext}
\usepackage{multirow}
\usepackage{times}
\usepackage{epsfig}
\usepackage{amsfonts}
\usepackage{amssymb}
\usepackage{color}

\def\tsc#1{\csdef{#1}{\textsc{\lowercase{#1}}\xspace}}
\tsc{WGM}
\tsc{QE}
\tsc{EP}
\tsc{PMS}
\tsc{BEC}
\tsc{DE}

\begin{document}
\shorttitle{}
\shortauthors{}

\title [mode = title]{DiverGAN: An Efficient and Effective Single-Stage Framework for Diverse Text-to-Image Generation} 

\author{Zhenxing Zhang}

\credit{Conceptualization, Data curation, Formal analysis, Investigation, Methodology, Software, Validation, Visualization, Writing - original draft}

\address{}

\author{Lambert Schomaker}

\credit{Conceptualization, Formal analysis, Methodology, Funding acquisition, Resources, Project administration, Supervision, Writing - review $\&$ editing}

\begin{abstract}
In this paper, we concentrate on the text-to-image synthesis task that aims at automatically producing perceptually realistic pictures from text descriptions. 
Recently, several single-stage methods have been proposed to deal with the problems of a more complicated multi-stage modular architecture. However, they often suffer from the lack-of-diversity issue, yielding similar outputs given a single textual sequence. 
To this end, we present an efficient and effective single-stage framework (DiverGAN) to generate diverse, plausible and semantically consistent images according to a natural-language description.
DiverGAN adopts two novel word-level attention modules, i.e., a channel-attention module (CAM) and a pixel-attention module (PAM), which model the importance of each word in the given sentence while allowing the network to assign larger weights to the significant channels and pixels semantically aligning with the salient words.
After that, Conditional Adaptive Instance-Layer Normalization (CAdaILN) is introduced to enable the linguistic cues from the sentence embedding to flexibly manipulate the amount of change in shape and texture, further improving visual-semantic representation and helping stabilize the training.
Also, a dual-residual structure is developed to preserve more original visual features while allowing for deeper networks, resulting in faster convergence speed and more vivid details. 
Furthermore, we propose to plug a fully-connected layer into the pipeline to address the lack-of-diversity problem, since we observe that a dense layer will remarkably enhance the generative capability of the network, balancing the trade-off between a low-dimensional random latent code contributing to variants and modulation modules that use high-dimensional and textual contexts to strength feature maps. Inserting a linear layer after the second residual block achieves the best variety and quality.
Both qualitative and quantitative results on benchmark data sets demonstrate the superiority of our DiverGAN for realizing diversity, without harming quality and semantic consistency. 
\end{abstract}


\maketitle

\section{Introduction}
The goal of text-to-image synthesis is to automatically yield perceptually plausible pictures, given textual descriptions. Recently, this topic rapidly gained attention in computer-vision and natural-language processing communities due to its extensive range of potential real-world applications including art creation, computer-aid design, data augmentation for training image classifiers, photo-editing, the education of young children, etc. Nevertheless, text-to-image generation is still an extremely challenging cross-modal task, since it not only requires a thorough semantic understanding of natural-language descriptions, but also requires a conversion of textual-context features into high-resolution images. The common paradigm involves a deep generative model implementing the cross-domain information fusion. 
\begin{figure*}
  \begin{minipage}[b]{1.0\linewidth}
  \centerline{\includegraphics[width=180mm]{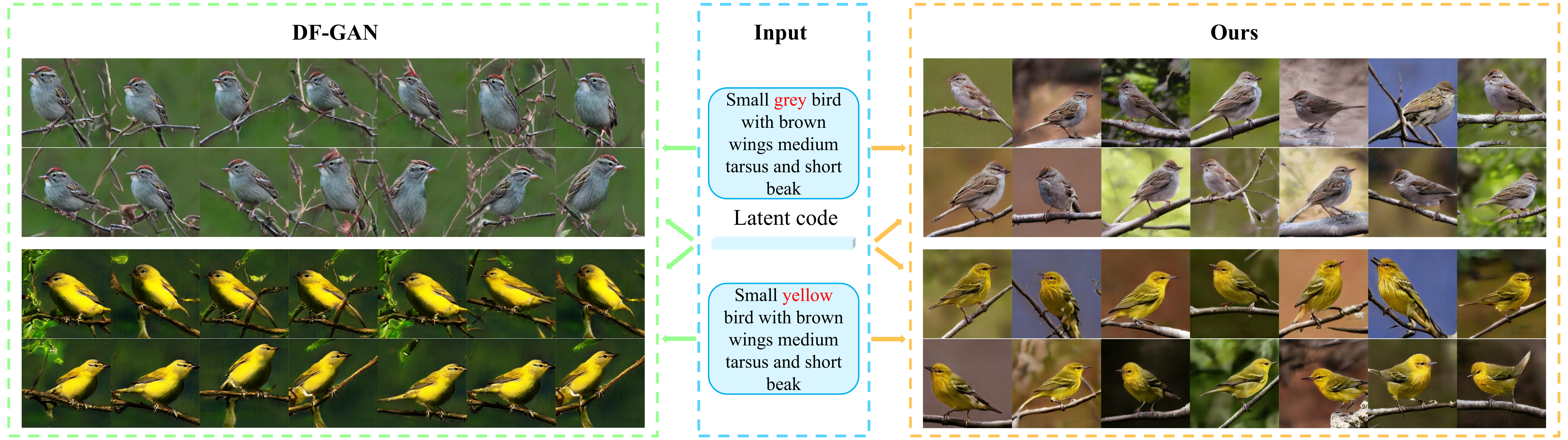}}
  \end{minipage}
  \caption{Diversity comparison between DF-GAN \cite{tao2020df} and our DiverGAN based on a single text description at the input. DF-GAN (Left) tends to suppress the random latent code, synthesizing slight variations of a bird. We propose DiverGAN (Right), an efficient and effective framework that is able to avoid the lack-of-diversity issue and yield diverse and high-quality pictures.}
  \vspace{-0.1in}
  \label{com} 
\end{figure*}

Thanks to the recent advances in the conditional generative adversarial network (CGAN) \cite{mirza2014conditional}, current text-to-image approaches have made tremendous progress in image quality and semantic consistency when given natural-language descriptions as inputs. 
Existing text-to-image approaches can be roughly cast into two types. The first category adopts a multi-stage modular architecture \cite{zhang2017stackgan, zhang2018stackgan++, xu2018attngan, li2019controllable, zhu2019dm, yin2019semantics, qiao2019mirrorgan} constructed with multiple generators and corresponding discriminators, producing visually realistic samples in a coarse-to-fine manner. 
Simultaneously, the output of the initial-stage network is fed into a next-stage network to generate an image with a higher resolution. 
While such a text-to-image procedure has been proven to be useful for the generation task, it requires training of several networks and is thus time-consuming \cite{tao2020df}. Even worse, the convergence and stability of the final generator network heavily relies on the earlier generators \cite{zhang2020dtgan}. 

To address the issues mentioned above, a second category of text-to-image synthesis methods has been recently studied in \cite{tao2020df, zhang2020dtgan}, merely leveraging a generator/discriminator pair to produce photo-realistic images which semantically align with corresponding textual descriptions. 
Thanks to the power of deep generative neural networks, feature fusion and generation process can be integrated into one single-stage procedure. 

However, the methods in this second category achieve structural simplicity and superior performance, but there  still  exist two  significant issues. Firstly, these approaches suffer from the mode-collapse problem \cite{arjovsky2017wasserstein}, in which the generator derails and synthesizes an inappropriate image for the text input. In less unfortunate cases of mode collapse, the CGAN generates a set of very similar output images, conditioned on the single natural-language description at the input. As shown in Fig.~\ref{com}, DF-GAN \cite{tao2020df} fails to produce {\em diverse} samples, although noise is present on the input.
This serious obstacle will drastically degrade the {\em diversity} of generated images, limiting their applicability in practice. For instance in data augmentation, robust
classification can only be achieved if a wide range of shapes was generated.
Secondly, existing single-stage approaches modulate the feature map by just adopting a global sentence vector, which lacks detailed and fine-grained information at the word level, and prevents the model from manipulating parts of the image generation according to natural-language descriptions and qualifications \cite{xu2018attngan, li2019controllable}.

One possible explanation for the lack-of-diversity issue is that the single-stage generator focuses more on the textual-context information that is high-dimensional and structured but ignores the random latent code which is responsible for variations \cite{mao2019mode}.
Considering the fact that single-stage methods usually utilize modulation modules to reinforce the visual feature map for each scale in order to ensure image quality and semantic consistency, the conditional contexts are likely to provide stronger control over the output image than the latent code, and thus, the generator yields nearly identical instances from a single text description.

 In this paper, we propose to develop an efficient and effective single-stage framework (DiverGAN) for yielding diverse and visually plausible instances that correspond well to the textual contexts. 
DiverGAN consists of four novel components discussed as follows.
 
Firstly, in order to stabilize the learning of the CGAN and boost the visual-semantic embedding in the visual feature map, Conditional Adaptive Instance-Layer Normalization (CAdaILN) is introduced to normalize the feature map in the layer and channel while employing the detailed and fine-grained linguistic cues to scale and shift the normalized feature map. Moreover, CAdaILN can help with flexibly controlling the amount of change in shape and texture, allowing for deeper networks and complementing the following modulation modules. 
For example, a good generator should be able to respond to the requirements specified by textual qualifiers for size ('large') and/or color ('red').

Secondly, 
we design two new types of word-level attention modules, i.e., a channel-attention module (CAM) and a pixel-attention module (PAM). These modules 
capture the semantic affinities between word-context vectors and feature maps in the channels and in the (2D) spatial dimensions. CAM and PAM not only guide the model to focus more on the crucial channels and pixels that are semantically correlated with the prominent words (e.g., adjectives and nouns) in the given textual description, but also alleviate the impact of semantically irrelevant and redundant information. More importantly, with CAM and PAM, DiverGAN can effectively disentangle the attributes of the text description while accurately controlling the regions of synthetic images.

Thirdly, we present a dual-residual block constructed with two residual modules, each of which contains convolutional layers, CAdaILN, exploiting a ReLU activation function followed by a modulation module. The dual-residual block not only benefits the CGAN convergence by retaining more original visual features, but also efficiently improves network capacity, resulting in high-quality images, with more details.

The text-to-image pipeline built upon the above three ingredients is enough to be capable of producing perceptually realistic images semantically matching with the natural-language descriptions but still suffers from the lack-of-diversity problem.

Therefore, as the fourth remedy we propose, on the basis of a variety of experiments, to use a linear layer which significantly boosts the generative ability of the network. This dense layer forces the generator to explore a wider range of modes in the original data distribution. In other words, plugging a linear layer into the single-stage architecture will improve the control of a random latent code over the visual feature map, balancing the trade-off between a random latent code contributing to diversity and modulation modules that modulate the feature map based on word-context vectors. Simultaneously, the experimental results will indicate that inserting a linear layer after the second dual-residual block of the architecture achieves the best performance on visual quality and image diversity. As illustrated in Fig.~\ref{com}, our DiverGAN equipped with a fully-connected layer is capable of generating birds with different visual appearances of footholds, background colors, orientations and shapes on the CUB bird data set \cite{wah2011caltech}.

We perform comprehensive experiments on three benchmark data sets, i.e., Oxford-102 \cite{nilsback2008automated}, CUB bird \cite{wah2011caltech} and MS COCO \cite{lin2014microsoft}. 
Both quantitative and qualitative results demonstrate that the proposed DiverGAN has the capacity to synthesize impressively better images than current single-stage and multi-stage models including StackGAN++ \cite{zhang2018stackgan++}, MSGAN \cite{mao2019mode}, SDGAN \cite{yin2019semantics}, DMGAN \cite{zhu2019dm}, DF-GAN \cite{tao2020df}, DTGAN \cite{zhang2020dtgan}. The contributions of this work can be summarized as follows:

$\bullet$ We establish a novel single-stage architecture for the text-to-image synthesis task. Our framework mitigates the lack-of-diversity issue, producing diverse and high-resolution samples that are semantically correlated with textural descriptions.   

$\bullet$ CAdaILN is introduced to stabilize training as well as help modulation modules flexibly control the amount of change in shape and texture. 

$\bullet$ Two new types of word-level attention modules are designed to reinforce visual feature maps with word-context vectors. 

$\bullet$ To the best of our knowledge, we are the first to embed one fully-connected layer into the single-stage pipeline to deal with the lack-of-diversity issue in text-to-image generation.

The remainder of this paper is organized as follows. In Section~\ref{rw}, we review related works. Section~\ref{p} briefly introduces the basic theories of a GAN, a CGAN, mode collapse and dot-product attention. In Section~\ref{pa}, we describe the proposed DiverGAN in detail. Section~\ref{er} reports the experimental results and the paper is summarized in Section~\ref{c}.

\section{Related works}
\label{rw}
In this section, research fields correlated with our work are described, including overcoming mode collapse in the GAN and the CGAN, CGAN-based text-to-image synthesis and attention mechanisms. 
\subsection{Avoiding mode collapse}
Mode collapse is a common but crucial obstacle existing in the GAN.
There are two major directions to mitigate the lack-of-diversity and mode-collapse issues in the generative adversarial network (GAN) \cite{goodfellow2014generative}.
Some publications suggest adapting the objective strategy to optimize the discriminator and the training process, e.g., by employing a minibatch discriminator \cite{salimans2016improved}, spectral regularization \cite{liu2019spectral} or unrolled optimization \cite{metz2016unrolled}. 
Metz et al. \cite{metz2016unrolled} proposed Unrolled GAN introducing a new objective to make the generator be updated through an unrolled optimization of the discriminator.
Salimans et al. \cite{salimans2016improved} developed a minibatch discriminator to check multiple output samples in a minibatch and a novel generator objective to tackle the overtraining on the discriminator.
Liu et al. \cite{liu2019spectral} recently proposed the spectral regularization for handling mode collapse. They suggest that the optimization of the discriminator is associated with the spectral distributions of the weight matrix.

Still other papers suggest the use of auxiliary structures into a GAN to encourage the generator to explore more modes of the true data distribution, including an autoencoder \cite{che2016mode, zhao2016energy, berthelot2017began, larsen2016autoencoding}, multiple generators \textcolor{red}{\cite{ghosh2018multi, cong2020discrete}}, a conditional augmentation technique \cite{zhang2017stackgan}, etc. 
Che et al. \cite{che2016mode} presented a mode regularized GAN (ModeGAN) 
incorporating the autoencoder into the standard GAN to avoid the mode missing problem. Motivated by ModeGAN, VEEGAN \cite{srivastava2017veegan} built a reconstruction net to map the true data distribution to the latent codes so that the generator network is able to synthesize all the data modes. EBGAN \cite{zhao2016energy}, BEGAN \cite{berthelot2017began} and VAEGAN \cite{larsen2016autoencoding} also made efforts towards combining the GAN with the autoencoder. To combat the mode-collapse issue, Ghosh et al. \cite{ghosh2018multi} provided a multi-agent GAN architecture (MAD-GAN), where multiple generators are employed to capture different modes and one discriminator is designed to identify generated samples.

There are also several works for mitigating the lack-of-diversity problem in the CGAN. Mao et al. \cite{mao2019mode} presented a regularization term to encourage the generator to explore more minor modes and force the discriminator to concentrate on the instances from the minor modes. 
To produce diverse instances from the textual description, Zhang et al \cite{zhang2017stackgan} developed a conditional augmentation technique (CA) concatenating a latent code and a textual variable sampled from the conditional Gaussian distribution as the inputs of the generator. 

Nonetheless, these approaches either require exorbitant computational cost or are invalid for a single-stage architecture. Here, we take the multi-generator model \cite{ghosh2018multi} as the example. Assuming that various single-stage generators are exploited to learn all modes of the true data distribution to overcome mode collapse, we expect that different generators are able to fit diverse modes. However, in practice, the generators are still prone to falling into a singular mode for text-to-image synthesis, since each generator only pays attention to the same conditional context.  
We would like to propose to distinguish three levels of performance for a GAN
:

\medskip
\noindent
Level 1: Traditional mode collapse \cite{salimans2016improved, metz2016unrolled, arjovsky2017wasserstein, kodali2017convergence, srivastava2017veegan, che2016mode, zhao2016energy} - strange, inappropriate patterns become a point attractor in the non-linear cyclic process~\cite{dynsys2011} between the generator and the discriminator (cf. Figure 6 in~\cite{salimans2016improved});

\medskip
\noindent
Level 2: Light mode collapse \cite{mao2019mode, lee2018diverse, zhang2017stackgan, bang2021mggan, lin2018pacgan} - patterns from the training set become wholly or partially an attractor for the generator. This is akin to lookup-table behavior made possible by the high number of parameters in a GAN. This constitutes the {\em lack-of-diversity} problem (cf. Figure 1, left, this paper);

\medskip
\noindent
Level 3: Desired functionality - generation of diverse patterns that are semantically consistent with the text probe as regards the foreground and that provide a believable, natural pattern in the background.

\medskip 

This paper focuses on solving the lack-of-diversity issue, i.e.,  the {\em light mode collapse}, not the severe, traditional mode collapse. 

\subsection{CGAN in text-to-image generation}
Thanks to the critical improvements in generative approaches, especially the GAN and the conditional GAN or CGAN, inspiring advances in text-to-image generation have been made possible. According to the number of the generators and the discriminators these methods exploit, we roughly group them into two categories.

\textbf{Multi-stage.} Zhang et al. \cite{zhang2017stackgan, zhang2018stackgan++} presented StackGAN and StackGAN++ to synthesize considerably compelling instances from the textual descriptions in a coarse-to-fine way. 
Qiao et al. \cite{qiao2019mirrorgan} proposed MirrorGAN exploiting 
an image caption model to regenerate the text description from a fake sample in order to boost the semantic relevancy between textual contexts and visual contents. Zhu et al. \cite{zhu2019dm} developed DMGAN which applied a dynamic memory module to improve the quality of initial images. Yin et al. \cite{yin2019semantics} designed a Siamese structure and conditional Batch Normalization to implement semantic consistency of synthetic images. 
SegAttnGAN \cite{gou2020segattngan} proposed to employ additional segmentation information for image refinement process such that the network is capable of yielding images with realistic quality.

\textbf{Single-stage.} Reed et al. \cite{Radford2016UnsupervisedRL} are the first to use the single-stage CGAN to generate images from detailed text descriptions. However, the quality of synthetic images is limited due to the simple structure and immature training techniques of the CGAN. Tao et al. \cite{tao2020df} presented DF-GAN adopting a matching-aware zero-centered gradient penalty (MA-GP) loss to cope with the problems of the multi-stage architecture. Nonetheless, DF-GAN just made use of a fully-connected layer to merge the feature map and a sentence vector, lacking an efficient modulation mechanism. Zhang et al. \cite{zhang2020dtgan} proposed DTGAN that introduced two novel attention modules and conditional normalization to generate high-resolution and semantically consistent images. Despite its excellent performance, DTGAN still suffers from the lack-of-diversity issue, which will be shown in Section~\ref{fc}. 

\subsection{Attention mechanism}
One vital property of our human visual system is that humans are capable of focusing more on the salient parts of an image and ignoring unimportant regions. Inspired by this, attention mechanisms are invented to guide the network to concentrate on the most discriminative local features and filter out irrelevant information. Thanks to their advantages over traditional methods with respect to feature processing and feature selection, attention mechanisms have been extensively explored in a series of computer-vision fields, such as automatic segmentation \cite{mei2021automatic, tang2021dsunet}, image fusion \cite{fang2020cross}, object tracking \cite{zhang2021csart}, video deblurring \cite{zhang2020attention}, scene classification \cite{bi2021multi}, image super-resolution \cite{wang2020single} and action recognition \cite{zheng2020spatial}.

There have been attention ways in text-to-image synthesis, since attention mechanisms play an essential role in bridging the semantic gap between vision and language. On the one hand, Xu et al. \cite{xu2018attngan} utilized a spatial-attention mechanism to derive the relationship between the image subregions and the words in a sentence. The most relevant subregions to the words were particularly focused. On the other hand, Li et al. \cite{li2019controllable} designed a channel-wise attention mechanism on the basis of Xu et al. \cite{xu2018attngan}. 
However, the aforementioned works adopt the weighted sum of converted word features as the new feature map which is largely different from the original feature map. Moreover, they both equally treat all words playing different roles in generating samples. 

As opposed to them, we target to modulate per-scale feature map with word features across both channel and spatial dimensions, while also retaining the basic features to some extent for the purpose of improving image quality and stable the learning of the CGAN. In the meantime, we model the importance of each word to emphasize the salient words (e.g., adjectives and nouns) in the given text description. 
The experimental results conducted on three benchmark data sets validate the effectiveness of our proposed attention modules compared to the prior methods.
\section{Preliminaries}
\label{p}
\subsection{Generative adversarial network (GAN)}
A GAN comprises two nets: the generator network $G$ and the discriminator network $D$, which are perceived as playing a minmax zero-sum game. To be concrete, the aim of $G$ is to capture the distribution of real data while yielding plausible images to trick $D$, whereas $D$ is optimized to classify a sample as real or fake. Concurrently, $G$ and $D$ are typically implemented by deep neural networks. Mathematically, the minmax objective $V(G, D)$ for the GAN can be denoted as follows:
\begin{equation}
\begin{split}
\min_{G}\max_{D}V(G,D)=&\mathbb{E}_{x\sim p_{data}(x)}\left[\log D(x)\right ]+\\
&\mathbb{E}_{z\sim p_{z}(z)}\left[\log (1 - D(G(z))) \right ]
\end{split}
\end{equation}
where $p_{data}(x)$ and $p_{z}(z)$ represent the distributions of true data $x$ and the random latent code $z$, respectively. 

$G$ tries to minimize the objective $V$ during the minmax two-player game, while $D$ aims to maximize it. This minmax zero-sum game is finished when the distribution of produced samples entirely overlaps $p_{data}(x)$.
\subsection{Conditional generative adversarial network (CGAN)}
As an extension of the GAN, a CGAN takes the conditional contexts $c$ (e.g. class labels, text descriptions and low-resolution images) and the random code $z$ as the inputs of $G$, while also outputting the samples which are correlated with $c$. In the meantime, $D$ entails distinguishing the real pair $(x, c)$ from the fake pair $(G(z, c), c)$. Concretely, the objective $V(G, D)$ for the CGAN is formulated as:
\begin{equation}
\begin{split}
\min_{G}\max_{D}V(G,D&)=\mathbb{E}_{x\sim p_{data}(x)}\left[\log D(x, c)\right ]+\\
&\mathbb{E}_{z\sim p_{z}(z),c\sim p_{c}(c)}\left[\log (1 - D(G(z, c), c)) \right ]
\end{split}
\end{equation}
where $p_{c}(c)$ denotes the distribution of $c$. 

In the task of text-to-image synthesis, $c$ is the given textual description. Let $\{ (I_{i}, C_{i})\}_{i=1}^{n}$ denote a set of $n$ image-text pairs for training, where $I_{i}$ represents an image and $C_{i}=(c_{i}^{1}, c_{i}^{2}, ..., c_{i}^{k})$ indicates a suite of $k$ natural-language descriptions. The goal of $G$ is to produce a visually plausible and semantically consistent sample $\hat{I}_{i}$ according to a text description $c_{i}$ randomly picked from $C_{i}$, where $c_{i}= (w_{1}, w_{2}, ..., w_{m})$ contains $m$ words. At the same time, $D$ is trained to distinguish the real text-image pair $(I_{i}, c_{i})$ from the synthetic text-image pair $(\hat{I}_{i}, c_{i})$.

\subsection{Mode collapse}
Mode collapse is a phenomenon where the model fails to learn all the modes of the true data distribution, and thus generated samples lack diversity \cite{arjovsky2017wasserstein}. In addition, for the CGAN, the produced instances from a single-condition context seem identical \cite{mao2019mode}. 
One of the main reasons for the lack-of-diversity issue is that the generator merely visits a part of the real data distribution and misses a few modes, limited by the generative capability of the network \cite{HUA2020101}. Although the generator tries to map random latent codes into the original data distribution, such a map is not surjective \cite{li2021tackling}. When the designed model is not powerful, the network can only capture a few modes of the real data distribution and tends to map different inputs into these modes. 

\begin{figure*}
  \begin{minipage}[b]{1.0\linewidth}
  \centerline{\includegraphics[width=180mm]{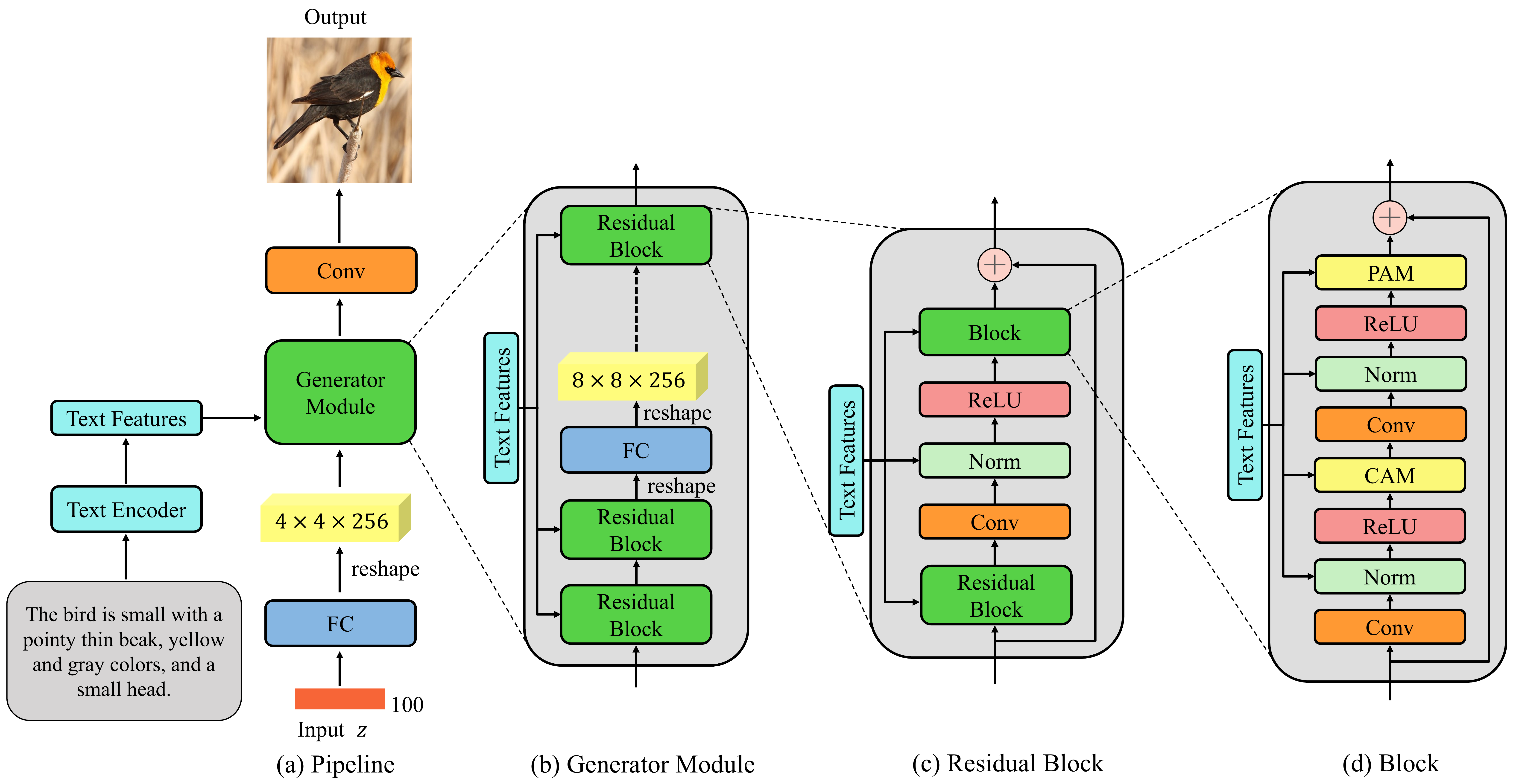}}
  \end{minipage}
  \caption{The overall architecture of the proposed DiverGAN. $z$ is a input latent code, FC is a fully-connected layer, Conv denotes a convolutional layer and Norm represents Conditional Adaptive Instance-Layer Normalization (CAdaILN) discussed in Section~\ref{Norm}. Additionally, CAM and PAM refer to the channel-attention module and pixel-attention module, respectively, discussed in Section~\ref{attention}. Moreover, the FC in (b) is employed to improve the generative ability of the generator, strengthening the control of $z$ over the visual feature map and boosting diversity. Note that we omit the up-sample operation between residual blocks in (b).}
  \vspace{-0.1in}
  \label{architecture} 
\end{figure*}
The lack-of-diversity issue becomes worse in the single-stage text-to-image pipeline. Generally, in the single-stage framework, a random latent code is taken as the input which is responsible for variations, while textual features serve as side information to modulate the visual feature map and determine the main visual contents. To boost the semantic consistency of generated instances, the generator usually leverages modulation modules for the per-scale feature map, which may make the network tend to concentrate on the textual-context features that are high-dimensional and structured, and ignore a low-dimensional latent code. In this work, we aim to enhance the generative ability of the network to reinforce the control of a random latent code over the visual feature map for the purpose of improving diversity.
\subsection{Scaled dot-product attention}
Generally, an attention function receives a query $q_{i}\in\mathbb{R}^{d}$ and a series of key-value pairs as the inputs and outputs a weighted sum of the values, where the attention weight on each value is calculated via a softmax function on the dot products of the query with all keys. The process of generating attention weights is formulated as follows:
\begin{equation}
Attention(q_{i},K,V)=Softmax(\frac{D(q_{i}, K^{T})}{\sqrt{d}})V
\end{equation}
where $D(\cdot)$ denotes the dot-product operation and $Softmax$ represents the softmax function. $K=(k_{1}, k_{2},...,k_{n}), k_{i}\in\mathbb{R}^{d}$, and $V=(v_{1}, v_{2},...,v_{n}), v_{i}\in\mathbb{R}^{d}$, refer to the keys and the values, respectively. $\frac{1}{\sqrt{d}}$ is the scaling factor.
\section{The proposed approach}
\label{pa}
In this section, 
we discuss the overall architecture of DiverGAN, depicted in Fig.~\ref{architecture}. After that, two novel types of attention modules, including a channel-attention module (CAM) and a pixel-attention module (PAM), are introduced to modulate the visual feature map with word-context embeddings. Subsequently, we describe the proposed Conditional Adaptive Instance-Layer Normalization (CAdaILN) leveraging the linguistic cues derived from the sentence vector to flexibly manipulate the amount of change in shape and texture. 
\subsection{Overall architecture}
Fig.~\ref{architecture} shows the overall architecture of DiverGAN. The first layer exploits a fully-connected layer to process a latent code $z\in \mathbb{R}^{100}$  and reshapes the result  as the initial feature map $F_{0}\in \mathbb{R}^{4 \times 4 \times 256}$. After that, $F_{0}$ is taken into the basic generator module (see Fig.~\ref{architecture}(b)) that mainly comprises seven dual-residual blocks (see Fig.~\ref{architecture}(c)) receiving textual embeddings derived from a text encoder as extra conditional contexts to strengthen the visual feature map. 

Fig.~\ref{architecture}(c) shows the details of our modified dual-residual module. 
Our designed dual-residual block consists of two residual modules (see Fig.~\ref{architecture}(d)), each of which is constructed with a set of convolutional layers, CAdaILN, ReLU activation functions followed with modulation modules (i.e., a channel-attention module (CAM) and a pixel-attention module (PAM)) taking the word-context features as side information to modulate the visual feature map. 
Furthermore, we plug a stack of a convolutional layer and CAdaILN activated by ReLU between two residual modules to relieve the control of contextual information. 
The dual-residual block not only accelerates convergence speed by retaining more original visual features than cascade convolutions, but also enables us to easily increase the depth of the network, efficiently improving network capacity and resulting in high-quality images with more details. 

To deal with the lack-of-diversity issue, we conduct a series of experiments on structure design and try to reinforce the control of the random latent code over the visual feature map to enhance variants. We observe that a dense layer can significantly boost the generative ability of the network, encouraging the model to explore minor modes of the true data distribution. For this reason, 
we attempt to insert one linear layer into our pipeline. The experimental results demonstrate that embedding a fully-connected layer after the second dual-residual block achieves appealing performance on visual quality and image diversity.
More specifically, the output feature map $F_{2}\in \mathbb{R}^{8 \times 8 \times 256}$ of the second dual-residual block is first resized to $\mathbb{R}^{16384}$ and put into a linear layer that maintains the dimension of the input features. Afterwards, $F_{2}$ is reshaped to $\mathbb{R}^{8 \times 8 \times 256}$ again and passed to the next dual-residual block. Then, the output of the basic generator module (see Fig.~\ref{architecture}(b)) is sent to one convolutional layer activated by tahn function to generate a final sample. 

\textbf{Why does inserting a dense layer address the lack-of-diversity issue?} By inserting a fully-connected layer, the network cannot exploit the spatial 2D layout of the preceding feature maps and needs to encode all the necessary information in a single 1D vector (embedding) as the basis for an unfolding in 2D by the later layers. As a result, we will have a representation at this point in the architecture that lends itself for injection of random noise with a subsequently increased diversity in the generated patterns: Because of the 1D bottleneck, the network cannot easily replicate (partial) 2D patterns from the early feature maps. This avoids the 'lookup-table' property that many GAN architectures have.

\textbf{Why does embedding a fully-connected layer after the second residual block achieve the best variety and quality?}  If the dense layer is too early in the network, it obtains early, crude featural information that is insufficient to generate semantically consistent output patterns. If you have an early feature map representing small bird components, the network will not be able to assemble a bird. On the other hand, if the dense layer is too late in the network, it will be fed by almost-complete patterns, and there will be a lack of diversity, since the system will operate as a lookup-table for the patterns in the training set that can then only be modified marginally by the last few layers. Additionally, it is extremely difficult to insert a dense layer after the third or later residual blocks due to the limited memory on GPUs. This can only be determined empirically.

To the best of our knowledge, we are the first to propose this kind of text-to-image architecture introducing one linear layer to improve the power of a random latent code to produce images of diverse modes. 
We expect that DiverGAN can provide a strong basis for the future developments of text-to-image generation. 
\subsection{Dual attention mechanism}
\label{attention}
It is well known that the semantic affinities between conditional contexts and visual feature maps are particularly critical for image synthesis. 
However, this correlation will become more complicated for text-to-image generation, since the given sentence contains a suite of words which have different contributions to synthetic samples. For instance, the adjective in the input text description will attend more to the produced sample than the definite article “the”.
Moreover, although our dual-residual structure is beneficial for model capacity and training stability, it may bring noise and redundant information. For these reasons, two new types of word-level attention modules, termed as a channel-attention module (CAM) and a pixel-attention module (PAM), are designed to explore the latent interplay between word-context features and visual feature maps.
CAM and PAM have the capacity to identify the significant words (e.g., adjectives and nouns) in the given text description and make the network assign more weights to the crucial channels and pixels semantically associated with these words.  
In addition, they can alleviate the effects of semantically irrelevant and redundant features from both channel and position perspectives.
\subsubsection{Channel-attention module (CAM)}
Channel maps have different responses to the words in the given sentence, and thus, the channels that respond to the prominent words in the sentence deserve more attention from the network. Here, we propose CAM to model the importance of each word while assigning larger weights to more useful channels semantically matching with the salient words. 

\begin{figure*}
  \begin{minipage}[b]{1.0\linewidth}
  \centerline{\includegraphics[width=180mm]{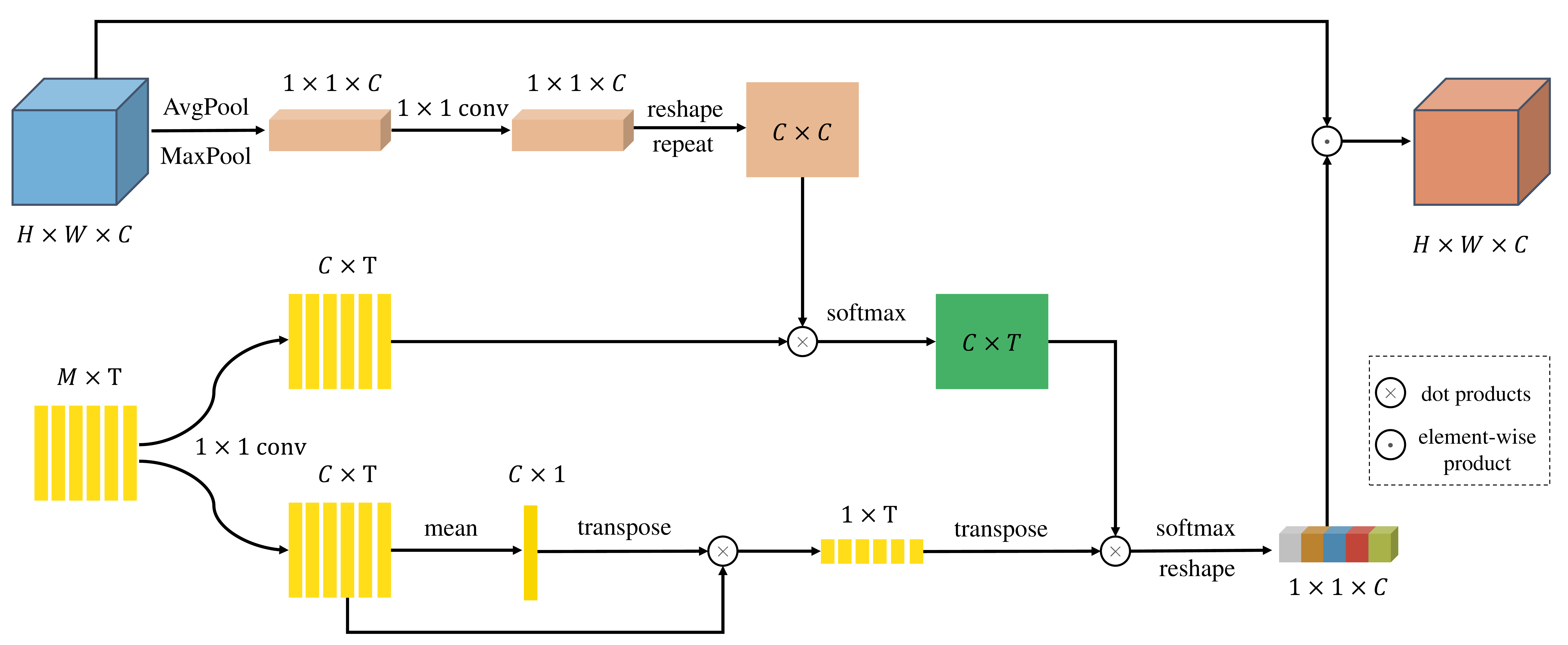}}
  \end{minipage}
  \caption{Overview of the proposed channel-attention module, which aims to assign larger weights to more useful channels semantically matching with the salient words. AvgPool and MaxPool refer to the global average pooling and max pooling, respectively. $H$, $W$ and $C$ denote the height, the width and the channel number of the visual feature map, respectively. $M$ is the dimension of the word embeddings and $T$ is the number of the words in the given text description. $1\times1$ conv indicates the $1\times1$ convolution operation.}
  \vspace{-0.1in}
  \label{CAM} 
\end{figure*}

Fig.~\ref{CAM} illustrates the detailed structure of CAM. Given a feature map $F_{c}\in\mathbb{R}^{H\times W\times C}$ (where $H$, $W$ and $C$ denote the height, the width and the channel number of $F_{c}$, respectively), we first adopt the global average pooling and max pooling to process it to aggregate holistic and discriminative information, thereby deriving two channel feature vectors $F_{ca}\in\mathbb{R}^{1\times 1\times C}$ and $F_{cm}\in\mathbb{R}^{1\times 1\times C}$. After that, $F_{ca}$ and $F_{cm}$ are fed into two different $1\times1$ convolution layers rather than one $1\times1$ convolution operation, since average pooling and max pooling acquire different globally spatial statistics \cite{li2021attention}. The outputs are then converted to $\mathbb{R}^{1\times C}$ and repeated $C$ times along dimension 1 to obtain two features: the average-pooling query $F_{caq}\in\mathbb{R}^{C\times C}$ and the max-pooling query $F_{cmq}\in\mathbb{R}^{C\times C}$, respectively. Mathematically,
\begin{align}
&F_{caq}=f_{re}(fc_{aq}(Avg(F_{c}))) \\
&F_{cmq}=f_{re}(fc_{mq}(Max(F_{c})))
\end{align}
where $Avg$ and $Max$ represent the global average pooling and max pooling, respectively. $fc_{aq}$ and $fc_{mq}$ denote $1\times1$ convolution layers and $f_{re}$ refers to the reshape and repeat operations. 

For word-context vectors $E\in \mathbb{R}^{M\times T}$ (where $M$ denotes the dimension of the word embeddings and $T$ denotes the number of the words in the given text description), we flow them into two different $1\times1$ convolution operations followed by ReLU activation to produce two contextual vectors: the key $F_{ck}\in \mathbb{R}^{C\times T}$ and the value $F_{cv}\in \mathbb{R}^{C\times T}$, which are in the common semantic space of the visual features. Next, we compute the mean of the value $F_{cv}$ along the dimension 2 and resize it into $\mathbb{R}^{1\times C}$. Meanwhile, we multiply the result and the value $F_{cv}$ to gain the contextual attention map $E_{ci}\in \mathbb{R}^{1\times T}$ that indicates the importance of each word in the sentence. Intuitively, a larger value in the attention map means that the corresponding word attends more to the synthetic image. The acquisition of the contextual attention map is formulated as:
\begin{align}
&F_{ck}=fc_{ck}(E) \\
&F_{cv}=fc_{cv}(E)\\
&E_{ci}=fc_{mean}(F_{cv})*F_{cv}
\end{align}
where $fc_{ck}$ and $fc_{cv}$ refer to $1\times1$ convolution layers followed with ReLU activation. $fc_{mean}$ represents the average and reshape operations. 

Afterwards, to model the semantic affinities between word-context features and channels, we conduct a dot-product operation between the queries and the key $F_{ck}\in \mathbb{R}^{C\times T}$, and apply a softmax function to get the contextually channel-wise atention matrix $CA\in \mathbb{R}^{C\times T}$ indicating the similarity weights between channels and words in the sentence. At last, the channel-attention weights $W_{c}$ are calculated through a softmax function on the dot products of $CA$ with the transpose of $E_{ci}$, and resized into $\mathbb{R}^{1\times 1\times C}$. We formulate a series of operations as:
\begin{align}
&CA_{a}=Softmax(D(F_{caq}, F_{ck})) \\
&CA_{m}=Softmax(D(F_{cmq}, F_{ck})) \\
&W_{ca}=f_{re}(Softmax(D(CA_{a}, E_{ci}^{T})))) \\
&W_{cm}=f_{re}(Softmax(D(CA_{m}, E_{ci}^{T}))))
\end{align}
where $D(\cdot)$ denotes the dot-product operation and $Softmax$ represents the softmax function. $CA_{a}\in \mathbb{R}^{C\times T}$ and $CA_{m}\in \mathbb{R}^{C\times T}$ refer to the contextually channel-wise attention matrixes of $F_{caq}$ and $F_{cmq}$, respectively. $f_{re}$ indicates the reshape operation. $W_{ca}\in \mathbb{R}^{1\times 1\times C}$ and $W_{cm}\in \mathbb{R}^{1\times 1\times C}$ are the channel-attention weights of $CA_{a}$ and $CA_{m}$, respectively.

After acquiring the attention scores of channels, we multiply them and the original feature map to re-weight the visual feature map. By doing so, the network will focus more on useful channels of the feature map and assign larger weights on them. At the same time, we design an adaptive gating method to dynamically merge the output feature maps of the global average pooling and max pooling. The process of getting the fused feature map $F_{cw}\in \mathbb{R}^{H\times W\times C}$ is described as: 
\begin{align}
&F_{caw}=W_{ca}\odot F_{c}\\
&F_{cmw}=W_{cm}\odot F_{c} \\
&g_{ca}=\sigma(fc_{ga}(f_{re}(F_{ca}))+fc_{gm}(f_{re}(F_{cm}))) \\
&F_{cw}=g_{ca}*F_{caw}+(1-g_{ca})*F_{cmw}
\end{align}
where $\odot$ is the element-wise multiplication. $F_{caw}$ and $F_{cmw}$ denote the rescaled feature maps. $g_{ca}$ represents the response gate for the fusion of visual feature maps. $f_{re}$ refers to the reshape operation that resizes $F_{ca}$ and $F_{cm}$  into $\mathbb{R}^{1\times C}$. $fc_{ga}$ and $fc_{gm}$ represent fully-connected layers that reduce the number of channels to 1 and $\sigma$ denotes the sigmoid function. 
\begin{figure*}
  \begin{minipage}[b]{1.0\linewidth}
  \centerline{\includegraphics[width=180mm]{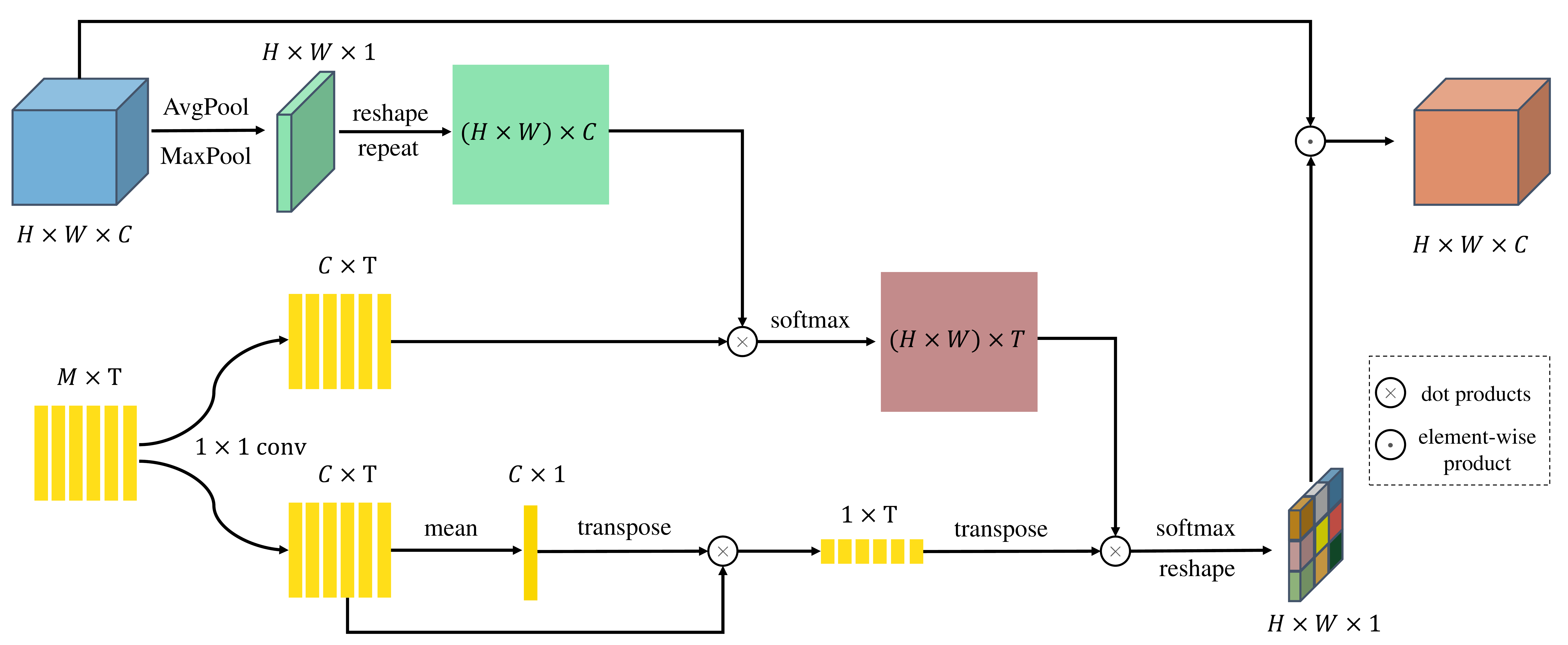}}
  \end{minipage}
  \caption{Overview of the proposed pixel-attention module, which aims to capture the spatial relationships between pixels and word embeddings, and allow significant pixels to acquire more weights. AvgPool and MaxPool denote the average pooling and max pooling in the spatial dimension, respectively. $H$, $W$ and $C$ denote the height, the width and the channel number of the visual feature map, respectively. $M$ is the dimension of the word embeddings and $T$ is the number of the words in the given text description. $1\times1$ conv indicates the $1\times1$ convolution operation.}
  \vspace{-0.1in}
  \label{PAM} 
\end{figure*}

To retain the basic features and stabilize the learning of the CGAN, we further apply an adaptive residual connection \cite{zhang2019self} to synthesize the final result $F_{cu}\in \mathbb{R}^{H\times W\times C}$. It is defined as follows:
\begin{equation}
F_{cu}=\gamma_{c}*F_{cw}+F_{c}
\end{equation}
where $\gamma_{c}$ is a learnable parameter which is initialized as 0.


\subsubsection{Pixel-attention module (PAM)}
As discussed above, CAM re-weights the visual feature map from the perspective of channel. However, in addition to the channels in the image feature map, the visual pixels are of central importance for the quality and semantic consistency. Hence, PAM is presented to effectively capture the spatial interplay between visual pixels and word-context vectors, allowing the significant pixels to gain more attentions from the network. Noteworthily, PAM is performed on the output of CAM, since the visual pixels in the same channel of the feature map still share the same weights. 

The framework of PAM is depicted in Fig.~\ref{PAM}. For a feature map $F_{s}\in\mathbb{R}^{H\times W\times C}$, we first perform the average pooling and max pooling in the spatial dimension over it to distill global features, acquiring two spatial feature vectors $F_{sa}\in\mathbb{R}^{H\times W\times 1}$ and $F_{sm}\in\mathbb{R}^{H\times W\times 1}$. Then, $F_{sa}$ and $F_{sm}$ are resized into $\mathbb{R}^{(H\times W)\times 1}$ and repeated $C$ times along dimension 2 to obtain two matrices: the average-pooling query $F_{saq}\in\mathbb{R}^{(H\times W)\times C}$ and the max-pooling query $F_{smq}\in\mathbb{R}^{(H\times W)\times C}$, respectively. The process of obtaining the queries is formulated as follows:
\begin{align}
&F_{saq}=f_{re}(Avg(F_{s})) \\
&F_{smq}=f_{re}(Max(F_{s}))
\end{align}
where $Avg$ and $Max$ denote the average pooling and max pooling in the spatial dimension, respectively. $f_{re}$ refers to the reshape and repeat operations. 

Given word-context features $E\in \mathbb{R}^{M\times T}$, we employ the same way as CAM to process them, producing the key $F_{sk}\in \mathbb{R}^{C\times T}$, the value $F_{sv}\in \mathbb{R}^{C\times T}$ and the contextual attention map $E_{si}\in \mathbb{R}^{1\times T}$. Specifically, 
\begin{align}
&F_{sk}=fc_{sk}(E) \\
&F_{sv}=fc_{sv}(E)\\
&E_{si}=fc_{mean}(F_{sv})*F_{sv}
\end{align}
where $fc_{sk}$ and $fc_{sv}$ refer to $1\times1$ convolution layers followed with ReLU activation. $fc_{mean}$ represents the average and reshape operations. 

After that, the spatial-semantic attention map $SA\in \mathbb{R}^{(H\times W)\times T}$ is achieved via a softmax function on the dot products of the queries with the key $F_{sk}\in \mathbb{R}^{C\times T}$, indicating the similarity weights between visual pixels and words in the sentence. Subsequently, we conduct a dot-product operation between $SA$ and the transpose of $E_{si}$, and leverage a softmax function to obtain the pixel-wise attention weights that are converted to $\mathbb{R}^{H\times W\times 1}$. The acquisition of the pixel-wise attention weights $W_{s}\in \mathbb{R}^{H\times W\times 1}$ is denoted as follows:
\begin{align}
&SA_{a}=Softmax(D(F_{saq}, F_{sk})) \\
&SA_{m}=Softmax(D(F_{smq}, F_{sk})) \\
&W_{sa}=f_{re}(Softmax(D(SA_{a}, E_{si}^{T})))) \\
&W_{sm}=f_{re}(Softmax(D(SA_{m}, E_{si}^{T}))))
\end{align}
where $D(\cdot)$ denotes the dot-product operation and $Softmax$ represents the softmax function. $SA_{a}\in \mathbb{R}^{(H\times W)\times T}$ and $SA_{m}\in \mathbb{R}^{(H\times W)\times T}$ refer to the contextually spatial-attention maps of $F_{saq}$ and $F_{smq}$, respectively. $W_{sa}\in \mathbb{R}^{H\times W\times 1}$ and $W_{sm}\in \mathbb{R}^{H\times W\times 1}$ are the pixel-wise attention weights of $SA_{a}$ and $SA_{m}$, respectively.

Next, same as CAM, we perform a matrix multiplication between the pixel-wise attention scores and the original feature map to facilitate the visual feature map. Meanwhile, in order to maintain the features, we concatenate the rescaled feature maps of the average pooling and max pooling, and put the result into a $1\times 1$ convolution layer followed with a ReLU function to generate the merged feature map $F_{sw}\in \mathbb{R}^{H\times W\times C}$. Then, an adaptive residual connection is adopted to get the final output $F_{su}\in \mathbb{R}^{H\times W\times C}$. This process is described as:
\begin{align}
&F_{saw}=W_{sa}\odot F_{s}\\
&F_{smw}=W_{sm}\odot F_{s} \\
&F_{sw}=fc_{con}([F_{saw};F_{smw}])\\
&F_{su}=\gamma_{s}*F_{sw}+F_{s}
\end{align}
where $\odot$ is the element-wise multiplication. $F_{saw}$ and $F_{smw}$ denote the rescaled feature maps. $[;]$ refers to the concatenation operation along the channel dimension and $fc_{con}$ represents the $1\times 1$ convolutional operation followed with a ReLU function. $\gamma_{s}$ is a learnable parameter which is initialized as 0. 
\begin{figure}[t]
   \begin{minipage}[b]{1.0\linewidth}
   \centerline{\includegraphics[width=85mm]{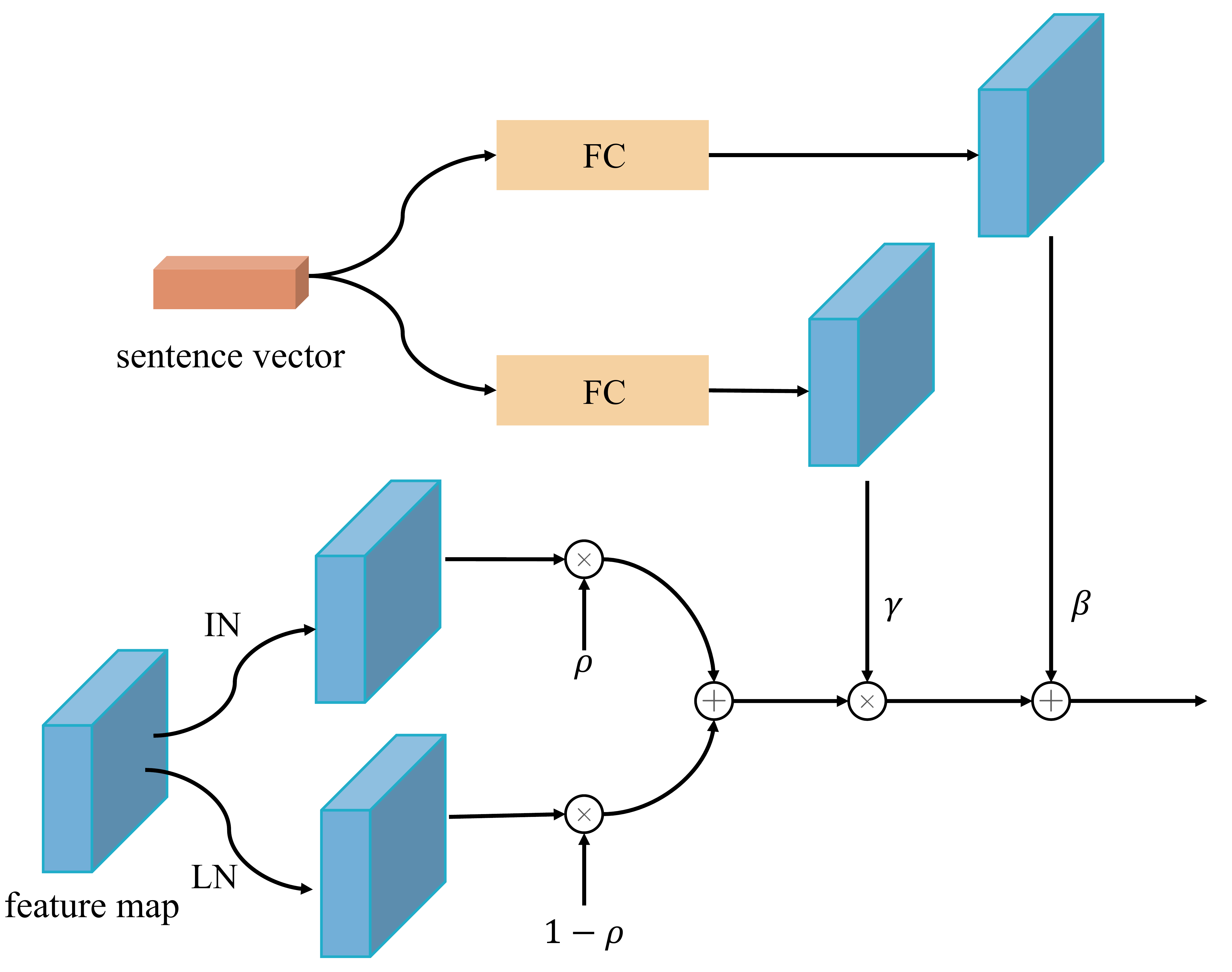}}
   \end{minipage}
   \caption{The process of Conditional Adaptive Instance-Layer Normalization (CAdaILN), which uses the linguistic cues derived from the sentence vector to flexibly control the amount of change in shape and texture. IN and LN denote Instance Normalization (IL) and Layer Normalization (LN), respectively. FC is a fully-connected layer. $\gamma$ and $\beta$ indicate sentence-level linguistic cues. $\rho$ is a learnable parameter that determines the ratio of IN and LN.}
   \vspace{-0.1in}
   \label{C1} 
\end{figure}

\textbf{Why do our attention modules work better?} Firstly, we model the importance of each word to emphasize the salient words (e.g., adjectives and nouns) in the given text description. This will make our generator concentrate more on the semantically related parts of the image. Secondly, in order to achieve better quality, we utilize average pooling and max pooling to acquire different globally spatial and channel information. Thirdly, an adaptive gating method is designed to dynamically merge the output feature maps of average pooling and max pooling, obtaining better performance. Fourthly, we specifically spread our attention weights to all the channels and pixels to enhance feature maps in several (!) layers, while applying an adaptive residual connection to synthesize the final result, in order to retain the basic features and stabilize the learning of the CGAN. We have not seen this in literature. Fifthly, different from the previous methods, we leverage our attention modules to modulate per-scale feature map across both channel and spatial dimensions. Sixthly, our proposed CAdaILN can help with flexibly controlling the amount of change in shape and texture, complementing our attention modules.

\subsection{Conditional Adaptive Instance-Layer Normalization (CAdaILN)}
\label{Norm}
In order to stabilize the training of the GAN \cite{goodfellow2014generative}, most existing text-to-image generation models \cite{xu2018attngan, li2019controllable, zhu2019dm, qiao2019mirrorgan, yin2019semantics} employ Batch Normalization (BN) \cite{santurkar2018does} applying the normalization to a whole batch of generated images instead for single ones. However, the convergence of BN heavily depends on the size of a batch \cite{lian2019revisit}. Furthermore, the advantage of BN is not obvious for text-to-image generation, since each synthetic image is more pertinent to the given text description and the feature map itself. To this end, CAdaILN is designed to perform the normalization in the layer and channel of the feature map $f$ and modulate the normalized feature map $\hat{f}$ with the linguistic cues captured from the global sentence vector $s$, illustrated in Fig.~\ref{C1}. More concretely, we employ two fully-connected layers $W_{1}$ and $W_{2}$ to transform the sentence vector $s$ into the linguistic cues $\gamma \in \mathbb{R}^{1\times 1\times C}$ and $\beta  \in \mathbb{R}^{1\times 1\times C}$. Moreover, we normalize the visual feature map with Instance Normalization (IN) and Layer Normalization (LN). After that, the normalized feature map $\hat{f}$ is acquired via the adaptive sum of the IN output $\hat{a}_{I^{}}$ and the LN output $\hat{a}_{L^{}}$. Afterwards, we leverage $\gamma$ and $\beta$ to scale and shift $\hat{f}$. The process of CAdaILN is formulated as follows:
\begin{align}
&\gamma=W_{1}s,\beta=W_{2}s   \\
&\hat{f}=\rho\odot a\hat{}_{I^{}}+(1-\rho)\odot a\hat{}_{L^{}}\\
&\hat{a}=\gamma\odot\hat{f}+\beta 
\end{align}
where the ratio of IN and LN is dependent on a learnable parameter $\rho \in \mathbb{R}^{1\times 1\times C}$, whose value is constrained to the range of [0, 1]. Moreover, $\rho$ is updated together with generator parameters. 

Notice that our proposed dual-attention mechanisms and CAdaILN are all easy-to-implement methods, although they seem complicated. 
\section{Experiments}
\label{er}
In this section, to prove the effectiveness of the proposed DiverGAN in diversity and producing visually realistic and semantically consistent images, we perform a wealth of quantitative and qualitative evaluations on three benchmark data sets, i.e., Oxford-102 \cite{nilsback2008automated}, CUB bird \cite{wah2011caltech} and MS COCO \cite{lin2014microsoft}. To be specific, we clarify the details of experimental settings in Section~\ref{es}. After that, the proposed DiverGAN is compared to previous CGAN-based approaches for text-to-image generation in Section~\ref{cws}. Subsequently, we analyze the contributions from different components of our DiverGAN in Section 5.3. 
\subsection{Experimental settings}
\label{es}
\textbf{Datasets.} We evaluate DiverGAN on three extensively employed data sets, which are used by StackGAN \cite{zhang2017stackgan}, AttnGAN \cite{xu2018attngan},
MirrorGAN \cite{qiao2019mirrorgan}, ControlGAN \cite{li2019controllable}, 
SDGAN \cite{yin2019semantics}, DM-GAN \cite{zhu2019dm}, DF-GAN \cite{tao2020df}, etc. 

$\bullet$ \textbf{Oxford-102.} The Oxford-102 data set includes 5,878 and 2,311 images for training and testing, respectively. Each image is accompanied by 10 textual descriptions.

$\bullet$ \textbf{CUB bird.} The CUB data set consists of 11,788 images, where 8,855 images belong to the training set and the other 2,933 images belong to the test set. Each image contains 10 sentences for text descriptions. 

$\bullet$ \textbf{MS COCO.} The MS COCO data set is a more challenging data set comprising 123,287 images which are divided into 82,783 training images and 40,504 test images. Each picture has 5 human annotated captions.

\textbf{Implementation details.} For the text encoder, following the method of AttnGAN \cite{xu2018attngan}, we utilize a pretrained bidirectional Long Short-Term Memory network \cite{schuster1997bidirectional} to acquire the word embeddings and the global sentence vector, respectively. We adopt the losses in DTGAN \cite{zhang2020dtgan} owing to its superior results. We set the dimension of the latent code to 100. As for the training, we leverage the Adam optimizer ~\cite{Kingma2015AdamAM} with $\beta=(0.0, 0.9)$ to train our network. We also follow the two time-scale update rule (TTUR) \cite{heusel2017gans} and set the learning rates for the generator and the discriminator to 0.0001 and 0.0004, respectively. The batch size is set to 16. Our DiverGAN is implemented by PyTorch~\cite{paszke2019pytorch}. All the experiments are performed on a single NVIDIA Tesla V100 GPU (32 GB memory).
\begin{table}
\caption{The IS of prior approaches and our DiverGAN on the CUB and Oxford-102 data sets. The best scores are in bold.}
\begin{center}
\begin{tabular}{l c c}
\hline
Methods & CUB $\uparrow$ & Oxford-102 $\uparrow$ \\
\hline
GAN-INT-CLS \cite{10.5555/3045390.3045503}& 2.88$\pm$0.04 & 2.66$\pm$0.03\\
GAWWN \cite{reed2016learning}& 3.62$\pm$0.07 & $-$\\
StackGAN \cite{zhang2017stackgan}& 3.70$\pm$0.04& 3.20$\pm$0.01\\
StackGAN++ \cite{zhang2018stackgan++}& 4.04$\pm$0.05& 3.26$\pm$0.01\\
AttnGAN \cite{xu2018attngan}& 4.36$\pm$0.03& $-$\\
MirrorGAN \cite{qiao2019mirrorgan}& 4.56$\pm$0.05& $-$\\
ControlGAN \cite{li2019controllable}& 4.58$\pm$0.09& $-$\\
SDGAN \cite{yin2019semantics}& 4.67$\pm$0.09& $-$\\
SegAttnGAN \cite{gou2020segattngan} & 4.82$\pm$0.05 & 3.52$\pm$0.09\\
DM-GAN \cite{zhu2019dm}& 4.75$\pm$0.07& $-$\\
DF-GAN \cite{tao2020df}& 4.86$\pm$0.04& 3.71$\pm$0.06\\
DTGAN \cite{zhang2020dtgan}& 4.88$\pm$0.03& 3.77$\pm$0.06\\
\hline
\textbf{Ours} &  $\bm{4.98\pm0.06}$& $\bm{3.99\pm0.05}$\\
\hline
\end{tabular}
\end{center}
\label{qc1}
\vspace{-0.1in}
\end{table}
\begin{table}
\caption{The FID of StackGAN++ \cite{zhang2018stackgan++}, AttnGAN \cite{xu2018attngan}, DF-GAN \cite{tao2020df} and and our DiverGAN on the CUB and COCO data sets. The best results are in bold.}
\begin{center}
\begin{tabular}{l c c}
\hline
Methods & CUB $\downarrow$ & COCO $\downarrow$\\
\hline
StackGAN++ \cite{zhang2018stackgan++}& 26.07& 51.62\\
AttnGAN \cite{xu2018attngan}& 23.98& 35.49\\
DF-GAN \cite{tao2020df}& 19.24& 28.92\\
\hline
\textbf{Ours} &  \textbf{15.63}& \textbf{20.52}\\
\hline
\end{tabular}
\end{center}
\label{qc2}
\vspace{-0.15in}
\end{table}

\textbf{Evaluation metrics.} Our DiverGAN is evaluated by calculating three widely used metrics including an inception score (IS) \cite{salimans2016improved}, a Fr\'echet inception distance (FID) \cite{szegedy2016rethinking} score and a learned perceptual image patch similarity (LPIPS) \cite{zhang2018unreasonable} score. The first two are to measure the visual quality, and the last one is employed to assess the diversity of generated samples. 

$\bullet$ \textbf{IS}. The IS is acquired via the KL divergence between the conditional class distribution and the marginal class distribution. 
\begin{figure*}
   \begin{minipage}[b]{1.0\linewidth}
   \centerline{\includegraphics[width=180mm]{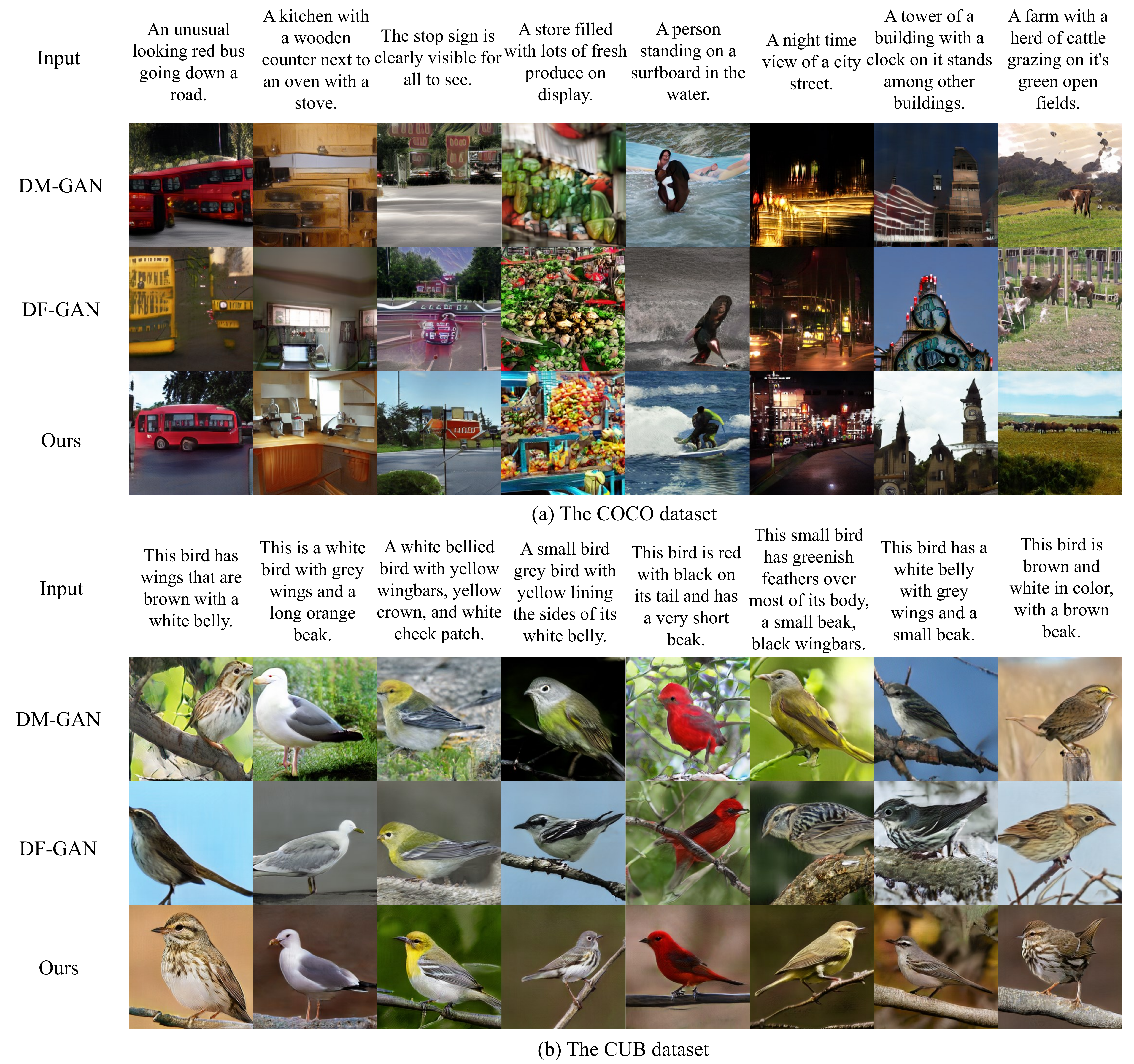}}
   \end{minipage}
   \caption{Qualitative comparison of DM-GAN \cite{zhu2019dm}, DF-GAN \cite{tao2020df} and our DiverGAN conditioned on the text descriptions on the COCO and CUB data sets.}
   \vspace{-0.1in}
  \label{Qualitative1} 
\end{figure*}
It’s defined as:
\begin{equation}
I=\text{exp}(\mathbb{E}_{x}[D_{KL}(p(y|x )\parallel p(y))])
\end{equation}
where $x$ is a generated sample and $y$ is the corresponding label obtained by a pre-trained Inception v3 network \cite{szegedy2016rethinking}. The produced samples are split into multiple groups and the IS is calculated on each group of images, then the average and standard deviation of the score are reported.
Higher IS demonstrates better quality among the generated images.

For the COCO data set, DTGAN \cite{zhang2020dtgan}, DFGAN \cite{tao2020df} and ObjGAN \cite{li2019object} argue that the IS fails to evaluate the synthetic samples and can be saturated, even over-fitted. Consequently, we do not compare the IS on the COCO data set. 

$\bullet$ \textbf{FID.} The FID computes the Fr\'echet distance between the distribution of generated samples and the distribution of true data. A lower FID means that the synthetic samples are closer to the corresponding real images. We use a pre-trained Inception v3 network to achieve the FID. 

For the Oxford-102 data set, we do not list the FID due to lack of compared scores. It should be noted that we synthesize 30000 pictures from unseen textual descriptions for the IS and FID. 

$\bullet$ \textbf{LPIPS.} LPIPS measures diversity by computing the average feature distance between synthetic images. The generated samples are meant to be diverse if the LPIPS score is large. The results of LPIPS will be discussed in Section~\ref{fc}

\subsection{Comparison with state-of-the-art CGAN-based methods}
\label{cws}
\begin{figure*}
   \begin{minipage}[b]{1.0\linewidth}
   \centerline{\includegraphics[width=180mm]{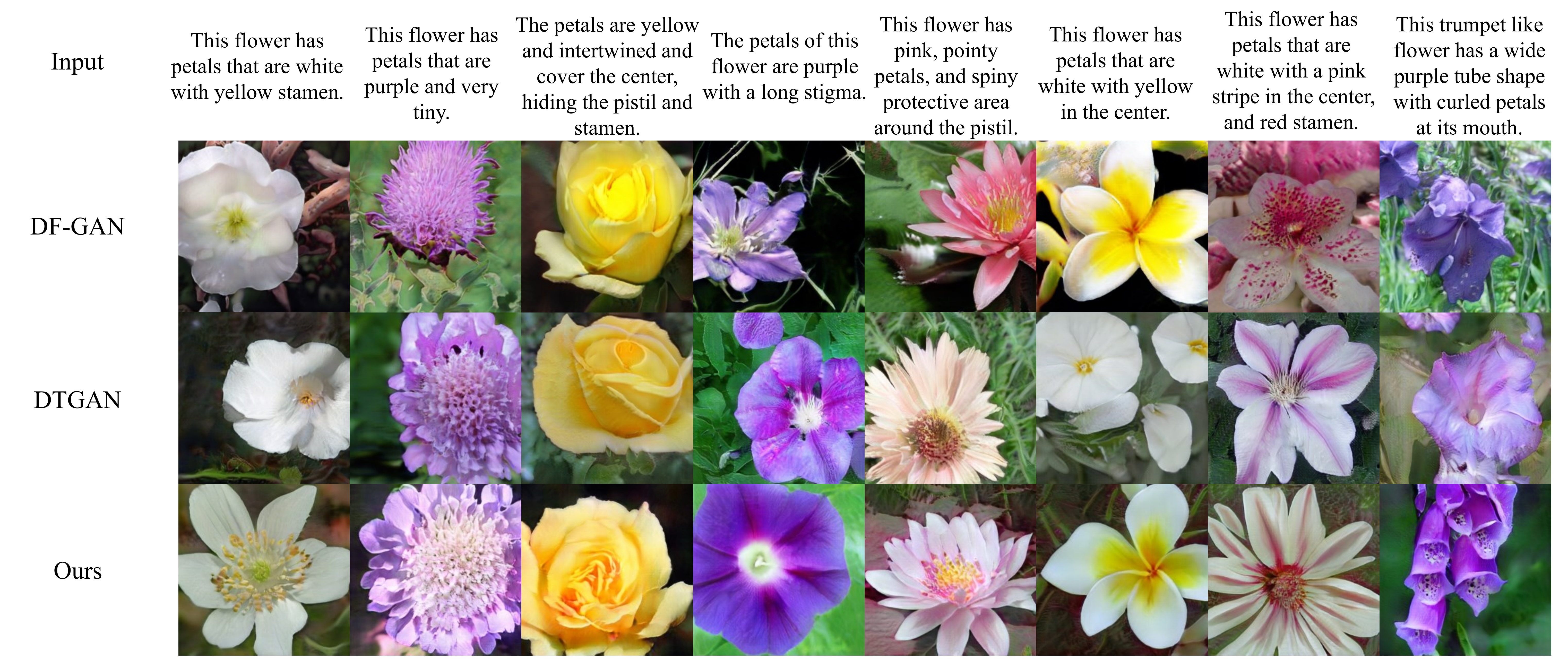}}
   \end{minipage}
   \caption{Qualitative comparison of DF-GAN \cite{tao2020df}, DTGAN \cite{zhang2020dtgan} and our DiverGAN conditioned on the text descriptions on the Oxford-102 data set.
    The bottom row does not show problems as in, e.g., DF-GAN 4th (shape), 7th (dull color),
    DT-GAN 1st (edges), 2nd (non-round shape) and 3rd (dull color).}
   \vspace{-0.1in}
  \label{Qualitative3} 
\end{figure*}
\begin{figure}[t]
  \begin{minipage}[b]{1.0\linewidth}
  \centerline{\includegraphics[width=85mm]{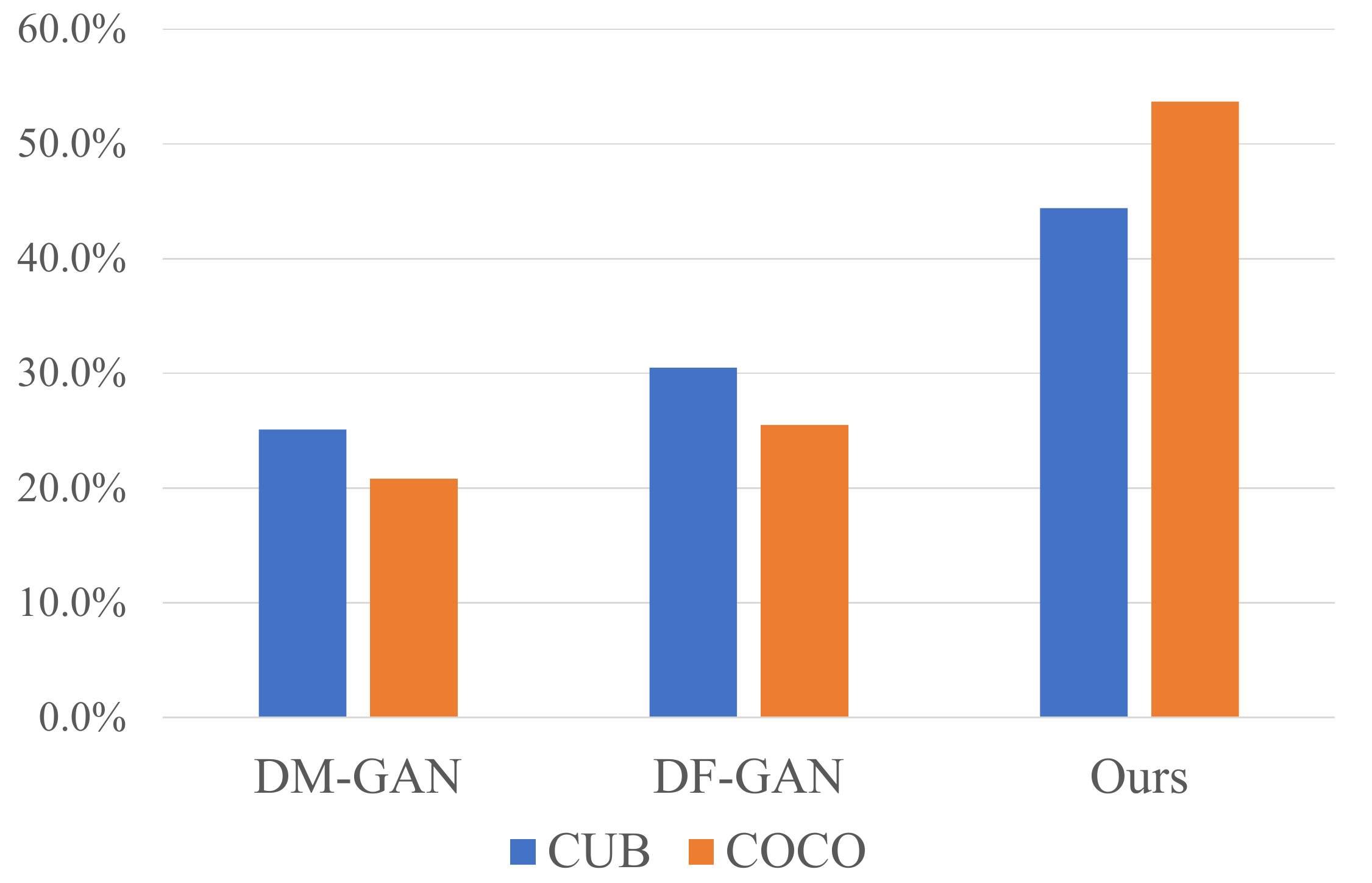}}
  \end{minipage}
  \caption{Human test results (ratio of 1st) of DM-GAN \cite{zhu2019dm}, DF-GAN  \cite{tao2020df} and our DiverGAN on the CUB and COCO data sets.}
  \vspace{-0.1in}
  \label{ht} 
\end{figure}
\subsubsection{Quantitative results}
We compare our method with previous single-stage \cite{tao2020df, zhang2020dtgan} and multi-stage \cite{zhang2017stackgan, zhang2018stackgan++, xu2018attngan, li2019controllable, zhu2019dm, yin2019semantics, qiao2019mirrorgan, gou2020segattngan} CGAN-based approaches on the CUB, Oxford-102 and COCO data sets. The IS of our DiverGAN and other compared models on the CUB and Oxford-102 data sets are reported in Table~\ref{qc1}. We can see that our DiverGAN achieves the best performance, significantly increasing the IS from 4.88 to 4.98 on the CUB data set and from 3.77 to 3.99 on the Oxford-102 data set. The experimental results indicate that our DiverGAN is capable of producing perceptually plausible pictures, with higher quality than state-of-the-art approaches. 

The comparison between our DiverGAN, StackGAN++ \cite{zhang2018stackgan++}, AttnGAN \cite{xu2018attngan} and DF-GAN \cite{tao2020df} with respect to the FID on the CUB and COCO data sets is shown in Table~\ref{qc2}. We can observe that our DiverGAN obtains a remarkably lower FID than compared methods on both data sets, which demonstrates that our generated data distribution is closer to the true data distribution. More specifically, we impressively reduce the FID from 28.92 to 20.52 on the challenging COCO data set and from 19.24 to 15.63 on the CUB data set.  

\subsubsection{Qualitative results}
In addition to quantitative comparison, we conduct qualitative experiments on the CUB, Oxford-102 and COCO data sets, which are illustrated in Fig.~\ref{Qualitative1} and Fig.~\ref{Qualitative3}.

Fig.~\ref{Qualitative1} shows the qualitative results of DM-GAN \cite{zhu2019dm}, DF-GAN  \cite{tao2020df} and our DiverGAN on the COCO and CUB data sets, indicating that our DiverGAN has the capacity to synthesize high-quality and semantic-consistency pictures conditioned on the text descriptions. For instance, in terms of complex scene generation, DiverGAN synthesizes a red bus with more vivid details than DF-GAN and DM-GAN in the column 1 of Fig.~\ref{Qualitative1}(a). It can also be seen that DiverGAN produces a kitchen with a plausible wooden counter ($2^{nd}$ column), a clear red sign on the road ($3^{rd}$ column), fresh vegetables and fruits with rich color distributions ($4^{th}$ column), a man surfing on the realistic sea waves ($5^{th}$ column) and a beautiful night street ($6^{th}$ column), whereas DM-GAN and DF-GAN both yield unclear objects ($1^{st}$, $2^{nd}$, $3^{rd}$, $7^{th}$ and $8^{th}$ column) and the background with a single color distribution ($4^{th}$, $5^{th}$ and $6^{th}$ column). More importantly, DiverGAN creates an impressive European clock tower in the column 7. The above results demonstrate that our DiverGAN equipped with the dual-residual structure is capable of capturing the crucial words in the sentence and highlighting the main objects of the image, generate a high-quality multi-object scene with vivid details. 
\begin{table*}[t]
\caption{Ablation study of our DiverGAN. CAM, PAM and FC represent the channel-attention module, the pixel-attention module and the insertion of the fully-connected layer between the first and the second residual block, respectively. The best results are in bold.}
\begin{center}
\scalebox{1}{
\begin{tabular}{c c c c c| c c c c}
\hline
\multirow{2}*{ID} & \multicolumn{4}{c|}{Components} & \multicolumn{2}{c}{CUB set} &  \multicolumn{1}{c}{\multirow{2}*{Oxford-102 (IS) $\uparrow$}} & \multicolumn{1}{c}{\multirow{2}*{COCO (FID) $\downarrow$}} \\ 
\cline{2-7}
 &  CAM  &  PAM  &  CAdaILN  &  FC & IS $\uparrow$ & FID $\downarrow$ & &\\  
\hline
M1 &  $\checkmark$ & $\checkmark$ & $\checkmark$ & $\checkmark$   &  $\bm{4.98\pm0.06}$ & \textbf{15.63} & $\bm{3.99\pm0.05}$ &  \textbf{20.52}\\ 

M2 &  $\checkmark$ & $\checkmark$ & $\checkmark$ & - & $4.91\pm0.06$ &  16.42 &$3.87\pm0.07$ & 22.53\\

M3 &  $\checkmark$ & $\checkmark$ & - & $\checkmark$   & $4.36\pm0.05$  &  24.17 & $3.46\pm0.04$ & 26.80\\

M4 &  $\checkmark$ & - & $\checkmark$ & $\checkmark$   &  $4.59\pm0.06$ & 20.86 & $3.77\pm0.04$ & 23.34\\

M5 &  - & $\checkmark$ & $\checkmark$ & $\checkmark$   & $4.73\pm0.04$  & 21.19 & $3.81\pm0.05$ & 23.40\\

\hline
\end{tabular}}
\end{center}
\label{as1}
\vspace{-0.15in}
\end{table*}

As can be observed in Fig.~\ref{Qualitative1}(b), DM-GAN, DF-GAN and DiverGAN all yield promising birds with consistent colors and shapes, but our method better concentrates on the semantically related parts of the image, synthesizing perceptually realistic birds. In addition, some backgrounds synthesized by DM-GAN ($1^{st}$, $2^{nd}$, $5^{th}$ and $7^{th}$ column) and DF-GAN ($5^{th}$, $7^{th}$ and $8^{th}$ column) are not 
plausible. It indicates that, with word-level attention modules, our model is able to bridge the gap between visual feature maps and word-context vectors, producing high-quality samples which semantically align with the text descriptions.
The qualitative comparison of DF-GAN, DTGAN \cite{zhang2020dtgan} and DiverGAN on the Oxford-102 is depicted in Fig.~\ref{Qualitative3}. We can observe that our approach synthesizes visually plausible flowers with more vivid details, richer color distributions and more clear shapes than DF-GAN and DTGAN, which confirms the effectiveness of DiverGAN.
\begin{table}
\caption{Ablation study on attention modules. Attn and ControlAttn indicate the attention modules in AttnGAN \cite{xu2018attngan} and ControlGAN \cite{li2019controllable}, respectively. CAM and PAM denote our channel-attention module and pixel-attention module, respectively. The best results are in bold.}
\begin{center}
\scalebox{0.87}{
\begin{tabular}{l c c c c c}
\hline
\multirow{2}*{Method} & \multicolumn{2}{c}{CUB set} &  \multicolumn{1}{c}{\multirow{2}*{Oxford-102 (IS) $\uparrow$}}\\ 
\cline{2-3}
& IS $\uparrow$ & FID $\downarrow$ & &\\  
\hline
Attn \cite{xu2018attngan} &  $4.52\pm0.04$ &  21.33 & $3.70\pm0.06$  \\
ControlAttn \cite{li2019controllable} & $4.63\pm0.05$ &  20.47 & $3.73\pm0.05$ \\
\hline
\textbf{CAM $\&$ PAM} & $\bm{4.91\pm0.06}$ & \textbf{16.42}  &$\bm{3.87\pm0.07}$ \\
\hline
\end{tabular}}
\end{center}
\label{as3}
\vspace{-0.1in}
\end{table}
\subsubsection{Human evaluation}
We perform a human test on the CUB and COCO data sets, so as to evaluate the image quality and the semantic consistency of DM-GAN \cite{zhu2019dm}, DF-GAN \cite{tao2020df} and our DiverGAN. We randomly select 100 images from both data sets, respectively. Given the same text descriptions, users are asked to choose the best sample synthesized by three approaches according to the image details and the corresponding natural-language description. In addition, the final scores are computed by two judges for fairness. As illustrated in Fig.~\ref{ht}, our method impressively outperforms DM-GAN and DF-GAN on both data sets, especially on the challenging COCO data set, which demonstrates the superiority of our proposed DiverGAN.
\begin{table*}
\caption{Ablation study on CAdaILN. BN-word indicates BN conditioned on the word vectors, BN-sent indicates Batch Normalization (BN) conditioned on the global sentence vector, CAdaILN-word indicates the CAdaILN function based on the word vectors and CAdaILN-sent indicates the CAdaILN function.}
\begin{center}
\begin{tabular}{c l| c c c c}
\hline
\multirow{2}*{ID} & \multirow{2}*{Architecture} & \multicolumn{2}{c}{CUB set} &  \multicolumn{1}{c}{\multirow{2}*{Oxford-102 (IS) $\uparrow$}} & \multicolumn{1}{c}{\multirow{2}*{COCO (FID) $\downarrow$}} \\ 
\cline{3-4}
 &  & IS $\uparrow$ & FID $\downarrow$ & &\\  
\hline
B1 &  Baseline & $4.36\pm0.05$  &  24.17 & $3.46\pm0.04$ & 26.80\\

B2 &  B1+BN-word & $4.60\pm0.04$  &  20.78 & $3.58\pm0.05$ & 24.58 \\

B3 &  B1+BN-sent & $4.65\pm0.05$  &  18.74 & $3.62\pm0.06$ & 25.64 \\

B4 &  B1+CAdaILN-word & $4.71\pm0.07$  &  17.19 & $3.73\pm0.05$ & 23.79\\

B5 &  B1+CAdaILN-sent & $\bm{4.91\pm0.06}$ & \textbf{16.42} &$\bm{3.87\pm0.07}$ & \textbf{22.53}\\
\hline
\end{tabular}
\end{center}
\label{as2}
\vspace{-0.1in}
\end{table*}
\begin{table}
\caption{Effect of the number of residual blocks in the dual-residual structure. Single-block and dual-block indicate the residual structure with a single block and two residual blocks, respectively. The best results are in bold.}
\begin{center}
\begin{tabular}{c| c c c}
\hline
\multicolumn{2}{c}{Metric}  & single-block & dual-block\\
\hline
\multirow{2}*{CUB} &
IS $\uparrow$ &   4.73$\pm$0.06 & $\bm{4.98\pm0.06}$ \\
& FID $\downarrow$ & 18.09 & \textbf{15.63} \\
\hline
\multicolumn{2}{c}{Oxford-102 (IS) $\uparrow$}  & 3.87$\pm$0.06 & $\bm{3.99\pm0.05}$\\
\hline
\end{tabular}
\end{center}
\label{as5}
\vspace{-0.1in}
\end{table}
\begin{figure*}
   \begin{minipage}[b]{1.0\linewidth}
   \centerline{\includegraphics[width=180mm]{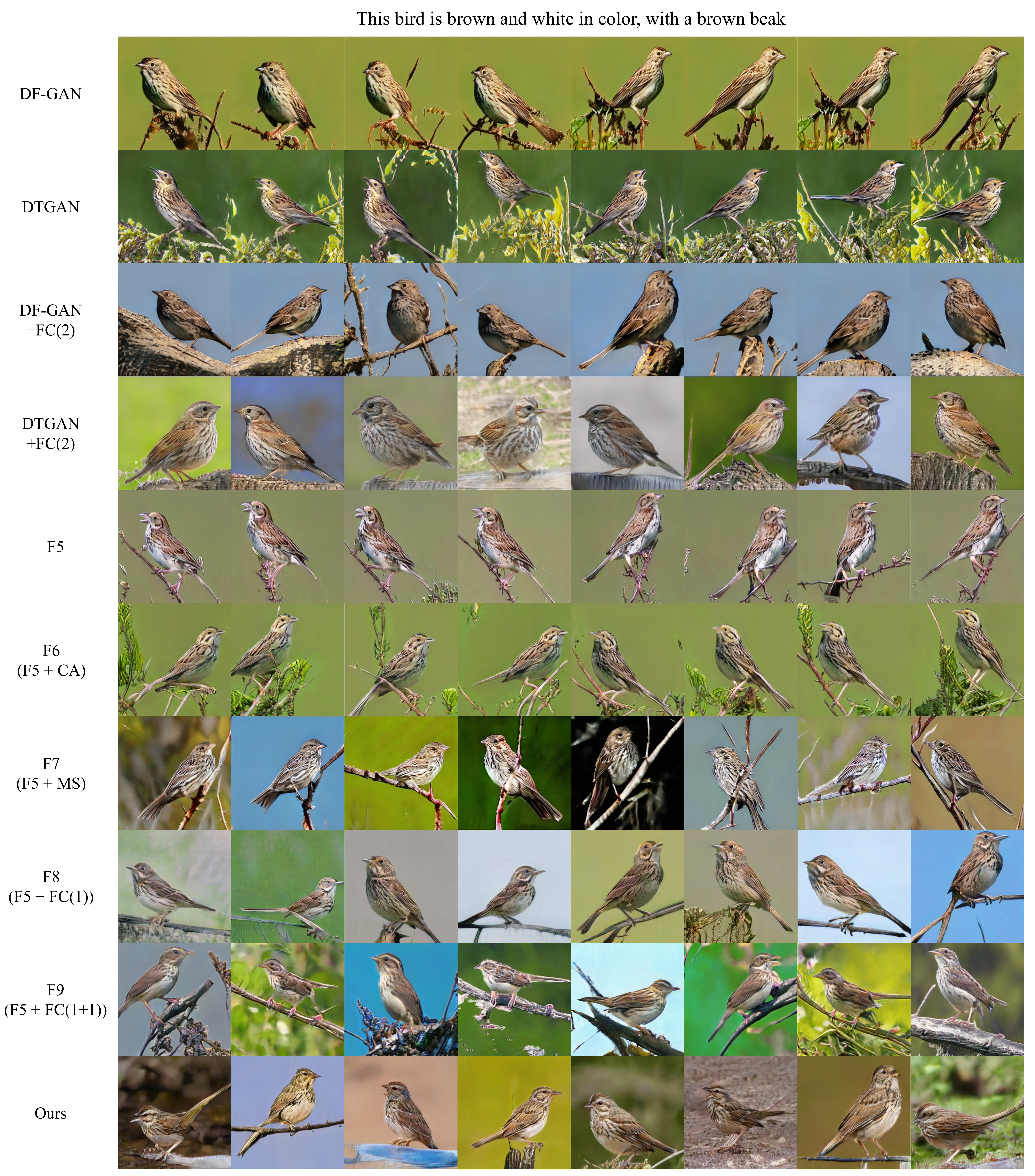}}
   \end{minipage}
   \caption{Qualitative results of other compared approaches and our DiverGAN (bottom row) when given a single text description, on the CUB data set. DF-GAN + FC(2) and DTGAN + FC(2) refer to DF-GAN and DTGAN that plug a dense layer after the second residual block, respectively. F5, F6, F7, F8 and F9 represent the corresponding models in Table~\ref{as4}.}
   \vspace{-0.1in}
  \label{Qualitative4} 
\end{figure*}
\subsection{Ablation studies of the proposed approach}
In order to evaluate the contributions from different components of our DiverGAN, we conduct extensive ablation studies on the CUB, Oxford-102 and COCO data sets. The novel components in our model include a channel-attention module (CAM), a pixel-attention module (PAM), CAdaILN, a dual-residual block and the insertion of a fully-connected layer (FC) between the first and the second residual block. We first quantitatively explore the effectiveness of each component by removing the corresponding part in DiverGAN step by step, i.e., M1: DiverGAN, M2: DiverGAN without the FC, M3: DiverGAN without CAdaILN, M4: DiverGAN without PAM, M5: DiverGAN without CAM. It is worth mentioning that we do not delete the dual-residual block in our ablation studies, since it is the basic structure in DiverGAN. All the results are reported in Table~\ref{as1}.
\begin{table*}
\caption{Effectiveness of the fully-connected layer (FC) in image diversity. Baseline (F5) corresponds to model M2 - DiverGAN removing the FC - in Table~\ref{as1}. CA and MS indicate conditioning augmentation (CA) and a mode-seeking regularization term (MS), respectively. FC(1) and FC(1+1) represent the insertion of one and two fully-connected layers after the first residual block, respectively. The best results are in bold.}
\begin{center}
\scalebox{1}{
\begin{tabular}{c l| c c c c c }
\hline
\multirow{2}*{ID} & \multirow{2}*{Architecture} & \multicolumn{3}{c}{CUB set} &  \multicolumn{1}{c}{\multirow{2}*{Oxford-102 (IS) $\uparrow$}} & \multicolumn{1}{c}{\multirow{2}*{COCO (FID) $\downarrow$}}\\ 
\cline{3-5}
 &  & IS $\uparrow$ & FID $\downarrow$ & LPIPS $\uparrow$ & &\\  
\hline
F1 &  StackGAN++ \cite{zhang2018stackgan++}& $4.04\pm0.05$ &  26.07&  0.362 &$3.26\pm0.01$ & 51.62  \\

F2 &  MSGAN \cite{mao2019mode}& - &  25.53&  0.373&-  & -   \\
F3 & DF-GAN \cite{tao2020df}& 4.86$\pm$0.04& 19.24 &0.537&3.71$\pm$0.06&28.92\\
F4 & DTGAN \cite{zhang2020dtgan}& 4.88$\pm$0.03& 16.35 &0.544& 3.77$\pm$0.06 & 23.61\\
\hline
F5 &  Baseline &$4.91\pm0.06$ & 16.42&0.549 &$3.87\pm0.07$ & 22.53\\

F6 &  F5+CA \cite{zhang2018stackgan++}& $4.72\pm0.06$ &  17.61&  0.535&$3.82\pm0.06$ & 25.23 \\

F7 &  F5+MS \cite{mao2019mode}& $4.57\pm0.05$ &  18.48&  0.667&$3.82\pm0.05$ & 24.44 \\
\hline
F8 &  F5+FC(1) & $4.76\pm0.05$ &  18.34&  0.579&$3.87\pm0.07$ & 22.16 \\

F9 & F5+FC(1+1) & $4.73\pm0.06$ &  18.68& 0.655 &$3.85\pm0.06$ & 22.55 \\


\textbf{F10}& \textbf{ours}  & $\bm{4.98\pm0.06}$ &  \textbf{15.63}&  \textbf{0.682}& $\bm{3.99\pm0.05}$ & \textbf{20.52} \\

\hline
\end{tabular}}
\end{center}
\label{as4}
\vspace{-0.1in}
\end{table*}

By comparing M1 (DiverGAN) with M2 (removing the FC), the introduction of the FC remarkably enhances the IS from 4.91 to 4.98 on the CUB data set and from 3.87 to 3.99 on the Oxford-102 data set, and reduces the FID from 16.42 to 15.63 on the CUB data set and from 22.53 to 20.52 on the challenging COCO data set. The experimental results demonstrate the importance of adopting a linear layer in DiverGAN. 
By exploiting CAdaILN in our DiverGAN, M1 performs better than M3 (removing CAdaILN), confirming the effectiveness of the proposed CAdaILN. To be specific, the IS is improved by 0.62 on the CUB data set and 0.53 on the Oxford-102 data set, and the FID is decreased by 8.54 on the CUB data set and 6.28 on the COCO data set. Furthermore, M1 achieves better results than both M4 (removing CAM) and M5 (removing PAM), which indicates that these two new types of word-level attention modules can help the generator yield more realistic images.
\subsubsection{Effectiveness of the attention modules}
To prove the effectiveness of our CAM and PAM, we also explore the performance of DiverGAN with other attention modules in text-to-image generation methods on the CUB and Oxford-102 data sets. Concretely, we replace the CAM and PAM in each residual block of M2 (DiverGAN removing the FC) with the attention modules in AttnGAN \cite{xu2018attngan} and ControlGAN \cite{li2019controllable}, respectively. Table~\ref{as3} displays the comparable results. It can be observed that our proposed attention mechanisms outperform AttnGAN and ControlGAN, improving the IS by 0.28 on the CUB data set and 0.14 on the Oxford-102, and reducing the FID from 20.47 to 16.42 on the CUB data set, which verifies the superiority of our proposed CAM and PAM. 

The reason behind this result may be that the prior attention modules directly convert semantic vectors to visual feature maps, adopting the weighted sum of converted word features as the new feature map, which is largely different from the original feature map. However, our attention modules aim to strengthen the visual feature map according to the contextual-semantic relevance while preserving the basic features to some extent. Additionally, CAM and PAM can emphasize the salient words in the given sentence instead of equally treating all words. 
Noteworthily, with word-level attention modules, our DiverGAN also has the ability to manipulate the parts of generated samples, which we detail in Section~\ref{fc}.
\subsubsection{Effectiveness of the proposed CAdaILN}
To further verify the benefits of CAdaILN, we conduct an ablation study on normalization functions. We first design a baseline model by removing CAdaILN from DiverGAN (M3). Then, we compare the variants of normalization layers. Note that BN conditioned on the global sentence vector (BN-sent) and BN conditioned on the word vectors (BN-word) are based on semantic-conditioned Batch Normalization in SDGAN \cite{yin2019semantics} and the CAdaILN function with the word vectors (CAdaILN-word) is achieved through the word-level normalization method in SDGAN. The results of the ablation study are shown in Table~\ref{as2}. It can be observed that by comparing B2 with B4 and B3 with B5, CAdaILN significantly outperforms the BN whether using the sentence-level linguistic cues or the word-level linguistic cues. Moreover, by comparing B5 with B4, CAdaILN with the global sentence vector performs better than CAdaILN-word by improving the IS from 4.71 to 4.91 on the CUB data set and from 3.73 to 3.87 on the Oxford-102 data set, and reducing the FID from 17.19 to 16.42 on the CUB data set and from 23.79 to 22.53 on the COCO data set, since sentence-level features may be easier to train in our generator network than word-level features. The above analysis demonstrates the effectiveness of our designed CAdaILN.
\subsubsection{Effect of the number of residual blocks in the residual-structure}
To evaluate the impact of different number of residual blocks in the dual-residual structure, we compare the performance of a residual block (single-block) and two residual blocks (dual-block) on the CUB and Oxford-102 data sets, as depicted in Table.~\ref{as5}. By comparison with single-residual block, the proposed dual-residual block enhances the IS by 0.25 on the CUB data set and 0.12 on the Oxford-102 data set, and decreases the FID by 2.16 on the CUB data set, which shows the effectiveness of the dual-residual structure.
\subsubsection{Effectiveness of the fully-connected layer in image diversity}
\label{fc}
To validate the effectiveness of the dense layer in diversity, we perform quantitative evaluation on the CUB, Oxford-102 and COCO data sets, and qualitative comparison on the CUB data set. 

\textbf{Quantitative results.} The quantitative comparison is divided into three groups according to the goals. The first group is designed to evaluate the quality and diversity of samples synthesized by single-stage methods (i.e., DFGAN and DTGAN); StackGAN++ and MSGAN, which propose conditioning augmentation (CA) and a mode seeking regularization term (MS) to enhance the image diversity, respectively. The second group aims to compare our proposed fully-connected (FC) method with the CA and MS. For fair comparisons, we take the model M2 (DiverGAN removing the FC) in Table~\ref{as1} as the baseline model (F5). F6 (F5 + CA) and F7 (F5 + MS) add the CA and MS to F5 for comparison, respectively. In the last group (i.e., F8 (F5 + FC(1)), F9 (F5 + FC(1+1)) and F10 (DiverGAN)), we verify the effectiveness of the different ways the linear layer is inserted into F5. F8 and F9 indicate the models that plug one and two dense layers after the first residual block of F5, respectively. 
\begin{figure}[t]
  \begin{minipage}[b]{1.0\linewidth}
  \centerline{\includegraphics[width=85mm]{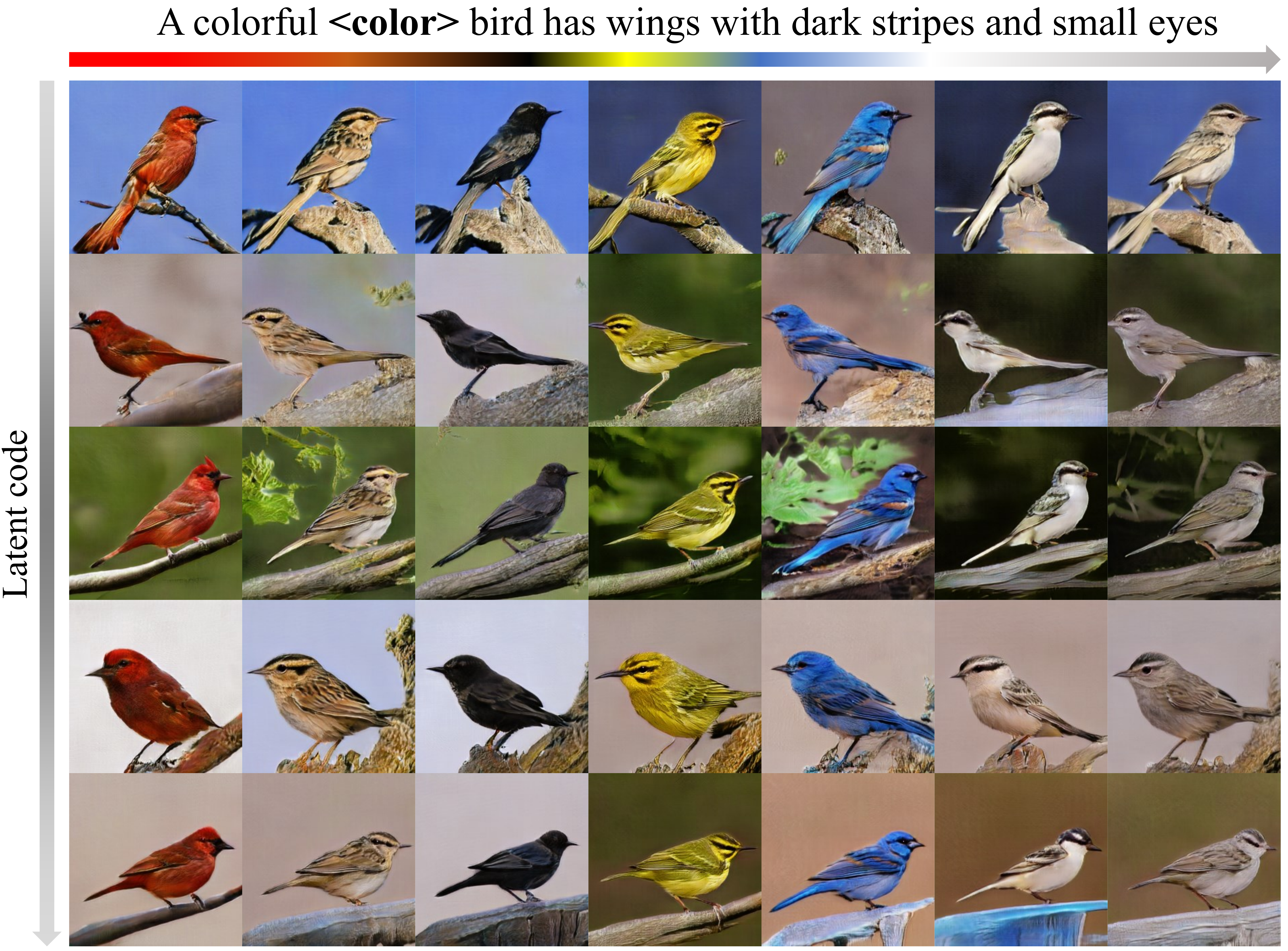}}
  \end{minipage}
  \caption{Generated samples of DiverGAN by changing the color attribute value in the input text description and the random latent codes, respectively.}
  \vspace{-0.1in}
  \label{img_ctrl} 
\end{figure}

The quantitative results are reported in Table~\ref{as4}. By comparing F10 (DiverGAN) with F1 (StackGAN++), F2 (MSGAN), F3 (DF-GAN) and F4 (DTGAN), our DiverGAN improves the LPIPS from 0.544 to 0.682 on the CUB data set, confirming the superiority of our DiverGAN in diversity. By comparison with F5, F7 enhances the LPIPS by 0.118 on the CUB data set, whereas it decreases the IS by 0.34 on the CUB data set and 0.05 on the Oxford-102 data set, and increases the FID by 2.06 on the CUB data set and 1.91 on the COCO data set. It demonstrates that, although the introduction of the MS boosts diversity, it may hurt image quality. F5 outperforms F6 showing CA may be not effective for current single-stage methods in diversity. F8, F9 and F10 all achieve larger LPIPS than F5, which validates the effectiveness of a linear layer in improving diversity. Furthermore, F10 performs better than F8 and F9 by improving the LPIPS from 0.655 to 0.682 and the IS from 4.76 to 4.98 on the CUB data set and 3.87 to 3.99 on the Oxford-102 data set, and reducing the FID by 2.71 on the CUB data set and 1.64 on the COCO data set, which indicates that inserting a linear layer after the second residual block of F5 obtains the best performance on quality and diversity. In conclusion, our DiverGAN is capable of just leveraging a generator/discriminator pair to generate perceptually realistic and diverse samples when given a single text description.   

\textbf{Qualitative results.} Subjective visual evaluation between DF-GAN, DTGAN, DF-GAN + FC(2), DTGAN + FC(2), F5, F6 (F5 + CA), F7 (F5 + MS), F8 (F5 + FC(1)), F9 (F5 + FC(1+1)) and F10 (DiverGAN) is presented in Fig.~\ref{Qualitative4}. DF-GAN + FC(2) and DTGAN + FC(2) refer to DF-GAN and DTGAN plugging a linear layer after the second residual block of the architecture, respectively.  

We can observe that, although DF-GAN ($1^{st}$ row) and DTGAN ($2^{nd}$ row) both perform well in quality, the shapes of the synthetic birds look similar and the background colors are the same. However, after inserting a dense layer into the framework, DF-GAN ($3^{rd}$ row) and DTGAN ($4^{th}$ row) both yield more diverse birds (e.g., different shapes, orientations and even background colors) than the original frameworks, demonstrating the generalizability of our proposed method. The reason behind similar backgrounds for DF-GAN + FC(2) may be that the background colors in DF-GAN only depend on the textual embeddings due to the introduction of the modulation module. By comparing our DiverGAN ($9^{th}$ row) with F6 (F5 + CA) and F7 (F5 + MS), we can see that a linear layer contributes to producing diverse birds with vivid details, whereas CA does not improve the diversity of birds and MS may affect the image quality. For example, in the row 9, the birds are on the branch or ground, and the orientations of birds, background colors and the visual appearances of footholds are different. Nonetheless, the birds in the row 5 still have similar shapes and the same background color, while the birds in the row 6 look a little blurry. 
By comparison with F5, we can observe that F8 (F5 + FC(1)), F9 (F5 + FC(1+1)) and DiverGAN all generate diverse samples, further confirming the effectiveness of the FC in diversity. In addition, DiverGAN generates realistic birds with higher quality than F6 and F7, which validate the superiority of DiverGAN.

To validate the sensitivity of our DiverGAN, we generate birds by modifying just one word in the given text description. As can be seen in Fig.~\ref{img_ctrl}, when we change the color attribute in the natural-language description, the proposed DiverGAN further produces semantically consistent birds according to the modified text while retaining visual appearances (e.g., shape, position and texture) correlated with the unmodified parts. Additionally, our method synthesizes a suite of birds with different visual appearances of footholds, background colors, orientations and shapes by changing latent codes. Therefore, with word-level attention modules and the FC method, DiverGAN is able to effectively disentangle attributes of the input text description while accurately controlling regions of the sample without hurting diversity. 


\section{Conclusion}
\label{c}
In this paper, we propose a unified, effective single-stage framework called DiverGAN for yielding diverse and perceptually realistic samples which are semantically related to given textual descriptions. DiverGAN exploits two new types of word-level attention modules, i.e., a channel-attention module (CAM) and a pixel-attention module (PAM), to make the generator concentrate more on useful channels and pixels that semantically match with the salient words in the natural-language description. In addition, Conditional Adaptive Instance-Layer Normalization (CAdaILN) is designed to adopt the linguistic cues to flexibly control the amount of change in shape and texture, strengthening visual-semantic representation and complementing modulation modules. Then, a dual-residual block is employed to accelerate convergence speed while enhancing image quality. Meanwhile, a fully-connected layer is introduced into our architecture to combat the lack-of-diversity problem by enhancing the generative ability of the network.
Extensive experiments on three benchmark data sets show that our DiverGAN achieves remarkably better performance than existing methods in quality and diversity. 
Furthermore, our presented components (i.e., CAM, PAM and CadaILN) are general methods, and can be readily integrated into current text-to-image architectures to reinforce feature maps with textual-context vectors. 
More importantly, our proposed pipeline tackles the lack-of-diversity issue existing in the current single-stage methods, and can serve as a strong basis for developing better single-stage models. 
For future work, we will investigate how to produce plausible samples which are semantically correlated with text descriptions in an unsupervised way and how to yield a suite of high-quality pictures based on regions of good solutions in the latent space.
One pervasive problem in the evaluation of the performance of image-generation methods is the degree of subjectivity involved. In case of human evaluations, more effort should be spent in guiding the human attention to aspects of the image: “Is the content semantically consistent with the text probe?” but also: “Is the background pattern believable/natural?”. Therefore, in future research, we will extend the questionnaire for the human subjects to ask more precisely what they think about the backgrounds of generated samples.

\bibliographystyle{model1-num-names}

\bibliography{cas-refs}

\begin{thebibliography}{56}
\expandafter\ifx\csname natexlab\endcsname\relax\def\natexlab#1{#1}\fi
\providecommand{\url}[1]{\texttt{#1}}
\providecommand{\href}[2]{#2}
\providecommand{\path}[1]{#1}
\providecommand{\DOIprefix}{doi:}
\providecommand{\ArXivprefix}{arXiv:}
\providecommand{\URLprefix}{URL: }
\providecommand{\Pubmedprefix}{pmid:}
\providecommand{\doi}[1]{\href{http://dx.doi.org/#1}{\path{#1}}}
\providecommand{\Pubmed}[1]{\href{pmid:#1}{\path{#1}}}
\providecommand{\bibinfo}[2]{#2}
\ifx\xfnm\relax \def\xfnm[#1]{\unskip,\space#1}\fi
\bibitem[{Tao et~al.(2020)Tao, Tang, Wu, Sebe, Wu, and Jing}]{tao2020df}
\bibinfo{author}{M.~Tao}, \bibinfo{author}{H.~Tang}, \bibinfo{author}{S.~Wu},
  \bibinfo{author}{N.~Sebe}, \bibinfo{author}{F.~Wu}, \bibinfo{author}{X.-Y.
  Jing},
\newblock \bibinfo{title}{{DF-GAN}: Deep fusion generative adversarial networks
  for text-to-image synthesis},
\newblock \bibinfo{journal}{arXiv preprint arXiv:2008.05865}
  (\bibinfo{year}{2020}).
\bibitem[{Mirza and Osindero(2014)}]{mirza2014conditional}
\bibinfo{author}{M.~Mirza}, \bibinfo{author}{S.~Osindero},
\newblock \bibinfo{title}{Conditional generative adversarial nets},
\newblock \bibinfo{journal}{arXiv preprint arXiv:1411.1784}
  (\bibinfo{year}{2014}).
\bibitem[{Zhang et~al.(2017)Zhang, Xu, Li, Zhang, Wang, Huang, and
  Metaxas}]{zhang2017stackgan}
\bibinfo{author}{H.~Zhang}, \bibinfo{author}{T.~Xu}, \bibinfo{author}{H.~Li},
  \bibinfo{author}{S.~Zhang}, \bibinfo{author}{X.~Wang},
  \bibinfo{author}{X.~Huang}, \bibinfo{author}{D.~N. Metaxas},
\newblock \bibinfo{title}{{StackGAN}: Text to photo-realistic image synthesis
  with stacked generative adversarial networks},
\newblock in: \bibinfo{booktitle}{Proceedings of the IEEE international
  conference on computer vision}, \bibinfo{year}{2017}, pp.
  \bibinfo{pages}{5907--5915}.
\bibitem[{Zhang et~al.(2018)Zhang, Xu, Li, Zhang, Wang, Huang, and
  Metaxas}]{zhang2018stackgan++}
\bibinfo{author}{H.~Zhang}, \bibinfo{author}{T.~Xu}, \bibinfo{author}{H.~Li},
  \bibinfo{author}{S.~Zhang}, \bibinfo{author}{X.~Wang},
  \bibinfo{author}{X.~Huang}, \bibinfo{author}{D.~N. Metaxas},
\newblock \bibinfo{title}{{StackGAN++}: Realistic image synthesis with stacked
  generative adversarial networks},
\newblock \bibinfo{journal}{IEEE transactions on pattern analysis and machine
  intelligence} \bibinfo{volume}{41} (\bibinfo{year}{2018})
  \bibinfo{pages}{1947--1962}.
\bibitem[{Xu et~al.(2018)Xu, Zhang, Huang, Zhang, Gan, Huang, and
  He}]{xu2018attngan}
\bibinfo{author}{T.~Xu}, \bibinfo{author}{P.~Zhang},
  \bibinfo{author}{Q.~Huang}, \bibinfo{author}{H.~Zhang},
  \bibinfo{author}{Z.~Gan}, \bibinfo{author}{X.~Huang},
  \bibinfo{author}{X.~He},
\newblock \bibinfo{title}{Attn{GAN}: Fine-grained text to image generation with
  attentional generative adversarial networks},
\newblock in: \bibinfo{booktitle}{Proceedings of the IEEE conference on
  computer vision and pattern recognition}, \bibinfo{year}{2018}, pp.
  \bibinfo{pages}{1316--1324}.
\bibitem[{Li et~al.(2019)Li, Qi, Lukasiewicz, and Torr}]{li2019controllable}
\bibinfo{author}{B.~Li}, \bibinfo{author}{X.~Qi},
  \bibinfo{author}{T.~Lukasiewicz}, \bibinfo{author}{P.~Torr},
\newblock \bibinfo{title}{Controllable text-to-image generation},
\newblock in: \bibinfo{booktitle}{Advances in Neural Information Processing
  Systems}, \bibinfo{year}{2019}, pp. \bibinfo{pages}{2065--2075}.
\bibitem[{Zhu et~al.(2019)Zhu, Pan, Chen, and Yang}]{zhu2019dm}
\bibinfo{author}{M.~Zhu}, \bibinfo{author}{P.~Pan}, \bibinfo{author}{W.~Chen},
  \bibinfo{author}{Y.~Yang},
\newblock \bibinfo{title}{{Dm-GAN}: Dynamic memory generative adversarial
  networks for text-to-image synthesis},
\newblock in: \bibinfo{booktitle}{Proceedings of the IEEE Conference on
  Computer Vision and Pattern Recognition}, \bibinfo{year}{2019}, pp.
  \bibinfo{pages}{5802--5810}.
\bibitem[{Yin et~al.(2019)Yin, Liu, Sheng, Yu, Wang, and
  Shao}]{yin2019semantics}
\bibinfo{author}{G.~Yin}, \bibinfo{author}{B.~Liu}, \bibinfo{author}{L.~Sheng},
  \bibinfo{author}{N.~Yu}, \bibinfo{author}{X.~Wang},
  \bibinfo{author}{J.~Shao},
\newblock \bibinfo{title}{Semantics disentangling for text-to-image
  generation},
\newblock in: \bibinfo{booktitle}{Proceedings of the IEEE Conference on
  Computer Vision and Pattern Recognition}, \bibinfo{year}{2019}, pp.
  \bibinfo{pages}{2327--2336}.
\bibitem[{Qiao et~al.(2019)Qiao, Zhang, Xu, and Tao}]{qiao2019mirrorgan}
\bibinfo{author}{T.~Qiao}, \bibinfo{author}{J.~Zhang}, \bibinfo{author}{D.~Xu},
  \bibinfo{author}{D.~Tao},
\newblock \bibinfo{title}{{MirrorGAN}: Learning text-to-image generation by
  redescription},
\newblock in: \bibinfo{booktitle}{Proceedings of the IEEE Conference on
  Computer Vision and Pattern Recognition}, \bibinfo{year}{2019}, pp.
  \bibinfo{pages}{1505--1514}.
\bibitem[{Zhang and Schomaker(2020)}]{zhang2020dtgan}
\bibinfo{author}{Z.~Zhang}, \bibinfo{author}{L.~Schomaker},
\newblock \bibinfo{title}{{DTGAN}: Dual attention generative adversarial
  networks for text-to-image generation},
\newblock \bibinfo{journal}{arXiv preprint arXiv:2011.02709}
  (\bibinfo{year}{2020}).
\bibitem[{Arjovsky et~al.(2017)Arjovsky, Chintala, and
  Bottou}]{arjovsky2017wasserstein}
\bibinfo{author}{M.~Arjovsky}, \bibinfo{author}{S.~Chintala},
  \bibinfo{author}{L.~Bottou},
\newblock \bibinfo{title}{Wasserstein generative adversarial networks},
\newblock in: \bibinfo{booktitle}{International conference on machine
  learning}, \bibinfo{organization}{PMLR}, \bibinfo{year}{2017}, pp.
  \bibinfo{pages}{214--223}.
\bibitem[{Mao et~al.(2019)Mao, Lee, Tseng, Ma, and Yang}]{mao2019mode}
\bibinfo{author}{Q.~Mao}, \bibinfo{author}{H.-Y. Lee}, \bibinfo{author}{H.-Y.
  Tseng}, \bibinfo{author}{S.~Ma}, \bibinfo{author}{M.-H. Yang},
\newblock \bibinfo{title}{Mode seeking generative adversarial networks for
  diverse image synthesis},
\newblock in: \bibinfo{booktitle}{Proceedings of the IEEE/CVF Conference on
  Computer Vision and Pattern Recognition}, \bibinfo{year}{2019}, pp.
  \bibinfo{pages}{1429--1437}.
\bibitem[{Wah et~al.(2011)Wah, Branson, Welinder, Perona, and
  Belongie}]{wah2011caltech}
\bibinfo{author}{C.~Wah}, \bibinfo{author}{S.~Branson},
  \bibinfo{author}{P.~Welinder}, \bibinfo{author}{P.~Perona},
  \bibinfo{author}{S.~Belongie}, \bibinfo{title}{{The Caltech-UCSD
  Birds-200-2011 Dataset}}, \bibinfo{type}{Technical Report}
  \bibinfo{number}{CNS-TR-2011-001}, California Institute of Technology,
  \bibinfo{year}{2011}.
\bibitem[{Nilsback and Zisserman(2008)}]{nilsback2008automated}
\bibinfo{author}{M.-E. Nilsback}, \bibinfo{author}{A.~Zisserman},
\newblock \bibinfo{title}{Automated flower classification over a large number
  of classes},
\newblock in: \bibinfo{booktitle}{2008 Sixth Indian Conference on Computer
  Vision, Graphics \& Image Processing}, \bibinfo{organization}{IEEE},
  \bibinfo{year}{2008}, pp. \bibinfo{pages}{722--729}.
\bibitem[{Lin et~al.(2014)Lin, Maire, Belongie, Hays, Perona, Ramanan,
  Doll{\'a}r, and Zitnick}]{lin2014microsoft}
\bibinfo{author}{T.-Y. Lin}, \bibinfo{author}{M.~Maire},
  \bibinfo{author}{S.~Belongie}, \bibinfo{author}{J.~Hays},
  \bibinfo{author}{P.~Perona}, \bibinfo{author}{D.~Ramanan},
  \bibinfo{author}{P.~Doll{\'a}r}, \bibinfo{author}{C.~L. Zitnick},
\newblock \bibinfo{title}{Microsoft coco: Common objects in context},
\newblock in: \bibinfo{booktitle}{European conference on computer vision},
  \bibinfo{organization}{Springer}, \bibinfo{year}{2014}, pp.
  \bibinfo{pages}{740--755}.
\bibitem[{Goodfellow et~al.(2014)Goodfellow, Pouget-Abadie, Mirza, Xu,
  Warde-Farley, Ozair, Courville, and Bengio}]{goodfellow2014generative}
\bibinfo{author}{I.~Goodfellow}, \bibinfo{author}{J.~Pouget-Abadie},
  \bibinfo{author}{M.~Mirza}, \bibinfo{author}{B.~Xu},
  \bibinfo{author}{D.~Warde-Farley}, \bibinfo{author}{S.~Ozair},
  \bibinfo{author}{A.~Courville}, \bibinfo{author}{Y.~Bengio},
\newblock \bibinfo{title}{Generative adversarial nets},
\newblock in: \bibinfo{booktitle}{Advances in neural information processing
  systems}, \bibinfo{year}{2014}, pp. \bibinfo{pages}{2672--2680}.
\bibitem[{Salimans et~al.(2016)Salimans, Goodfellow, Zaremba, Cheung, Radford,
  and Chen}]{salimans2016improved}
\bibinfo{author}{T.~Salimans}, \bibinfo{author}{I.~Goodfellow},
  \bibinfo{author}{W.~Zaremba}, \bibinfo{author}{V.~Cheung},
  \bibinfo{author}{A.~Radford}, \bibinfo{author}{X.~Chen},
\newblock \bibinfo{title}{Improved techniques for training {GAN}s},
\newblock in: \bibinfo{booktitle}{Advances in neural information processing
  systems}, \bibinfo{year}{2016}, pp. \bibinfo{pages}{2234--2242}.
\bibitem[{Liu et~al.(2019)Liu, Tang, Zhou, and Qiu}]{liu2019spectral}
\bibinfo{author}{K.~Liu}, \bibinfo{author}{W.~Tang}, \bibinfo{author}{F.~Zhou},
  \bibinfo{author}{G.~Qiu},
\newblock \bibinfo{title}{Spectral regularization for combating mode collapse
  in {GAN}s},
\newblock in: \bibinfo{booktitle}{Proceedings of the IEEE/CVF International
  Conference on Computer Vision}, \bibinfo{year}{2019}, pp.
  \bibinfo{pages}{6382--6390}.
\bibitem[{Metz et~al.(2016)Metz, Poole, Pfau, and
  Sohl-Dickstein}]{metz2016unrolled}
\bibinfo{author}{L.~Metz}, \bibinfo{author}{B.~Poole},
  \bibinfo{author}{D.~Pfau}, \bibinfo{author}{J.~Sohl-Dickstein},
\newblock \bibinfo{title}{Unrolled generative adversarial networks},
\newblock \bibinfo{journal}{arXiv preprint arXiv:1611.02163}
  (\bibinfo{year}{2016}).
\bibitem[{Che et~al.(2016)Che, Li, Jacob, Bengio, and Li}]{che2016mode}
\bibinfo{author}{T.~Che}, \bibinfo{author}{Y.~Li}, \bibinfo{author}{A.~P.
  Jacob}, \bibinfo{author}{Y.~Bengio}, \bibinfo{author}{W.~Li},
\newblock \bibinfo{title}{Mode regularized generative adversarial networks},
\newblock \bibinfo{journal}{arXiv preprint arXiv:1612.02136}
  (\bibinfo{year}{2016}).
\bibitem[{Zhao et~al.(2016)Zhao, Mathieu, and LeCun}]{zhao2016energy}
\bibinfo{author}{J.~Zhao}, \bibinfo{author}{M.~Mathieu},
  \bibinfo{author}{Y.~LeCun},
\newblock \bibinfo{title}{Energy-based generative adversarial network},
\newblock \bibinfo{journal}{arXiv preprint arXiv:1609.03126}
  (\bibinfo{year}{2016}).
\bibitem[{Berthelot et~al.(2017)Berthelot, Schumm, and
  Metz}]{berthelot2017began}
\bibinfo{author}{D.~Berthelot}, \bibinfo{author}{T.~Schumm},
  \bibinfo{author}{L.~Metz},
\newblock \bibinfo{title}{Be{GAN}: Boundary equilibrium generative adversarial
  networks},
\newblock \bibinfo{journal}{arXiv preprint arXiv:1703.10717}
  (\bibinfo{year}{2017}).
\bibitem[{Larsen et~al.(2016)Larsen, S{\o}nderby, Larochelle, and
  Winther}]{larsen2016autoencoding}
\bibinfo{author}{A.~B.~L. Larsen}, \bibinfo{author}{S.~K. S{\o}nderby},
  \bibinfo{author}{H.~Larochelle}, \bibinfo{author}{O.~Winther},
\newblock \bibinfo{title}{Autoencoding beyond pixels using a learned similarity
  metric},
\newblock in: \bibinfo{booktitle}{International conference on machine
  learning}, \bibinfo{organization}{PMLR}, \bibinfo{year}{2016}, pp.
  \bibinfo{pages}{1558--1566}.
\bibitem[{Ghosh et~al.(2018)Ghosh, Kulharia, Namboodiri, Torr, and
  Dokania}]{ghosh2018multi}
\bibinfo{author}{A.~Ghosh}, \bibinfo{author}{V.~Kulharia},
  \bibinfo{author}{V.~P. Namboodiri}, \bibinfo{author}{P.~H. Torr},
  \bibinfo{author}{P.~K. Dokania},
\newblock \bibinfo{title}{Multi-agent diverse generative adversarial networks},
\newblock in: \bibinfo{booktitle}{Proceedings of the IEEE conference on
  computer vision and pattern recognition}, \bibinfo{year}{2018}, pp.
  \bibinfo{pages}{8513--8521}.
\bibitem[{Cong et~al.(2020)Cong, Gui, Miao, Zhang, Wang, and
  Chen}]{cong2020discrete}
\bibinfo{author}{X.~Cong}, \bibinfo{author}{J.~Gui}, \bibinfo{author}{K.-C.
  Miao}, \bibinfo{author}{J.~Zhang}, \bibinfo{author}{B.~Wang},
  \bibinfo{author}{P.~Chen},
\newblock \bibinfo{title}{Discrete haze level dehazing network},
\newblock in: \bibinfo{booktitle}{Proceedings of the 28th ACM International
  Conference on Multimedia}, \bibinfo{year}{2020}, pp.
  \bibinfo{pages}{1828--1836}.
\bibitem[{Srivastava et~al.(2017)Srivastava, Valkov, Russell, Gutmann, and
  Sutton}]{srivastava2017veegan}
\bibinfo{author}{A.~Srivastava}, \bibinfo{author}{L.~Valkov},
  \bibinfo{author}{C.~Russell}, \bibinfo{author}{M.~U. Gutmann},
  \bibinfo{author}{C.~Sutton},
\newblock \bibinfo{title}{Vee{GAN}: Reducing mode collapse in {GAN}s using
  implicit variational learning},
\newblock \bibinfo{journal}{arXiv preprint arXiv:1705.07761}
  (\bibinfo{year}{2017}).
\bibitem[{Kodali et~al.(2017)Kodali, Abernethy, Hays, and
  Kira}]{kodali2017convergence}
\bibinfo{author}{N.~Kodali}, \bibinfo{author}{J.~Abernethy},
  \bibinfo{author}{J.~Hays}, \bibinfo{author}{Z.~Kira},
\newblock \bibinfo{title}{On convergence and stability of {GAN}s},
\newblock \bibinfo{journal}{arXiv preprint arXiv:1705.07215}
  (\bibinfo{year}{2017}).
\bibitem[{Broer and Takens(2011)}]{dynsys2011}
\bibinfo{author}{H.~Broer}, \bibinfo{author}{F.~Takens},
  \bibinfo{title}{Dynamical Systems and Chaos}, \bibinfo{publisher}{Springer},
  \bibinfo{address}{New York}, \bibinfo{year}{2011}.
\bibitem[{Lee et~al.(2018)Lee, Tseng, Huang, Singh, and Yang}]{lee2018diverse}
\bibinfo{author}{H.-Y. Lee}, \bibinfo{author}{H.-Y. Tseng},
  \bibinfo{author}{J.-B. Huang}, \bibinfo{author}{M.~Singh},
  \bibinfo{author}{M.-H. Yang},
\newblock \bibinfo{title}{Diverse image-to-image translation via disentangled
  representations},
\newblock in: \bibinfo{booktitle}{Proceedings of the European conference on
  computer vision (ECCV)}, \bibinfo{year}{2018}, pp. \bibinfo{pages}{35--51}.
\bibitem[{Bang and Shim(2021)}]{bang2021mggan}
\bibinfo{author}{D.~Bang}, \bibinfo{author}{H.~Shim},
\newblock \bibinfo{title}{{MGGAN}: Solving mode collapse using manifold-guided
  training},
\newblock in: \bibinfo{booktitle}{Proceedings of the IEEE/CVF International
  Conference on Computer Vision}, \bibinfo{year}{2021}, pp.
  \bibinfo{pages}{2347--2356}.
\bibitem[{Lin et~al.(2018)Lin, Khetan, Fanti, and Oh}]{lin2018pacgan}
\bibinfo{author}{Z.~Lin}, \bibinfo{author}{A.~Khetan},
  \bibinfo{author}{G.~Fanti}, \bibinfo{author}{S.~Oh},
\newblock \bibinfo{title}{Pac{GAN}: The power of two samples in generative
  adversarial networks},
\newblock \bibinfo{journal}{Advances in neural information processing systems}
  (\bibinfo{year}{2018}).
\bibitem[{Gou et~al.(2020)Gou, Wu, Li, Gong, and Han}]{gou2020segattngan}
\bibinfo{author}{Y.~Gou}, \bibinfo{author}{Q.~Wu}, \bibinfo{author}{M.~Li},
  \bibinfo{author}{B.~Gong}, \bibinfo{author}{M.~Han},
\newblock \bibinfo{title}{Segattn{GAN}: Text to image generation with
  segmentation attention},
\newblock \bibinfo{journal}{arXiv preprint arXiv:2005.12444}
  (\bibinfo{year}{2020}).
\bibitem[{Radford et~al.(2016)Radford, Metz, and
  Chintala}]{Radford2016UnsupervisedRL}
\bibinfo{author}{A.~Radford}, \bibinfo{author}{L.~Metz},
  \bibinfo{author}{S.~Chintala},
\newblock \bibinfo{title}{Unsupervised representation learning with deep
  convolutional generative adversarial networks},
\newblock \bibinfo{journal}{CoRR} \bibinfo{volume}{abs/1511.06434}
  (\bibinfo{year}{2016}).
\bibitem[{Mei et~al.(2021)Mei, Lei, Gu, Ye, Sun, Zhang, and
  Wang}]{mei2021automatic}
\bibinfo{author}{H.~Mei}, \bibinfo{author}{W.~Lei}, \bibinfo{author}{R.~Gu},
  \bibinfo{author}{S.~Ye}, \bibinfo{author}{Z.~Sun},
  \bibinfo{author}{S.~Zhang}, \bibinfo{author}{G.~Wang},
\newblock \bibinfo{title}{Automatic segmentation of gross target volume of
  nasopharynx cancer using ensemble of multiscale deep neural networks with
  spatial attention},
\newblock \bibinfo{journal}{Neurocomputing} \bibinfo{volume}{438}
  (\bibinfo{year}{2021}) \bibinfo{pages}{211--222}.
\bibitem[{Tang et~al.(2021)Tang, Zu, Hong, Yan, Peng, Xiao, Wu, Zhou, Zhou, and
  Wang}]{tang2021dsunet}
\bibinfo{author}{P.~Tang}, \bibinfo{author}{C.~Zu}, \bibinfo{author}{M.~Hong},
  \bibinfo{author}{R.~Yan}, \bibinfo{author}{X.~Peng},
  \bibinfo{author}{J.~Xiao}, \bibinfo{author}{X.~Wu},
  \bibinfo{author}{J.~Zhou}, \bibinfo{author}{L.~Zhou},
  \bibinfo{author}{Y.~Wang},
\newblock \bibinfo{title}{Da-dsunet: Dual attention-based dense su-net for
  automatic head-and-neck tumor segmentation in mri images},
\newblock \bibinfo{journal}{Neurocomputing} \bibinfo{volume}{435}
  (\bibinfo{year}{2021}) \bibinfo{pages}{103--113}.
\bibitem[{Fang et~al.(2020)Fang, Zhao, and Zhang}]{fang2020cross}
\bibinfo{author}{A.~Fang}, \bibinfo{author}{X.~Zhao},
  \bibinfo{author}{Y.~Zhang},
\newblock \bibinfo{title}{Cross-modal image fusion guided by subjective visual
  attention},
\newblock \bibinfo{journal}{Neurocomputing} \bibinfo{volume}{414}
  (\bibinfo{year}{2020}) \bibinfo{pages}{333--345}.
\bibitem[{Zhang et~al.(2021)Zhang, Zheng, Li, and Liu}]{zhang2021csart}
\bibinfo{author}{D.~Zhang}, \bibinfo{author}{Z.~Zheng},
  \bibinfo{author}{M.~Li}, \bibinfo{author}{R.~Liu},
\newblock \bibinfo{title}{Csart: Channel and spatial attention-guided residual
  learning for real-time object tracking},
\newblock \bibinfo{journal}{Neurocomputing} \bibinfo{volume}{436}
  (\bibinfo{year}{2021}) \bibinfo{pages}{260--272}.
\bibitem[{Zhang et~al.(2020)Zhang, Jiang, Wang, Huang, and
  Zhao}]{zhang2020attention}
\bibinfo{author}{X.~Zhang}, \bibinfo{author}{R.~Jiang},
  \bibinfo{author}{T.~Wang}, \bibinfo{author}{P.~Huang},
  \bibinfo{author}{L.~Zhao},
\newblock \bibinfo{title}{Attention-based interpolation network for video
  deblurring},
\newblock \bibinfo{journal}{Neurocomputing}  (\bibinfo{year}{2020}).
\bibitem[{Bi et~al.(2021)Bi, Zhang, and Qin}]{bi2021multi}
\bibinfo{author}{Q.~Bi}, \bibinfo{author}{H.~Zhang}, \bibinfo{author}{K.~Qin},
\newblock \bibinfo{title}{Multi-scale stacking attention pooling for remote
  sensing scene classification},
\newblock \bibinfo{journal}{Neurocomputing} \bibinfo{volume}{436}
  (\bibinfo{year}{2021}) \bibinfo{pages}{147--161}.
\bibitem[{Wang et~al.(2020)Wang, Lu, Li, Wang, Wang, and Chen}]{wang2020single}
\bibinfo{author}{Z.~Wang}, \bibinfo{author}{Y.~Lu}, \bibinfo{author}{W.~Li},
  \bibinfo{author}{S.~Wang}, \bibinfo{author}{X.~Wang},
  \bibinfo{author}{X.~Chen},
\newblock \bibinfo{title}{Single image super-resolution with attention-based
  densely connected module},
\newblock \bibinfo{journal}{Neurocomputing}  (\bibinfo{year}{2020}).
\bibitem[{Zheng et~al.(2020)Zheng, Zheng, Lu, and Wu}]{zheng2020spatial}
\bibinfo{author}{Y.~Zheng}, \bibinfo{author}{X.~Zheng},
  \bibinfo{author}{X.~Lu}, \bibinfo{author}{S.~Wu},
\newblock \bibinfo{title}{Spatial attention based visual semantic learning for
  action recognition in still images},
\newblock \bibinfo{journal}{Neurocomputing} \bibinfo{volume}{413}
  (\bibinfo{year}{2020}) \bibinfo{pages}{383--396}.
\bibitem[{Hua et~al.(2020)Hua, Rui, Cai, Wang, Zhang, and Wang}]{HUA2020101}
\bibinfo{author}{X.~Hua}, \bibinfo{author}{T.~Rui}, \bibinfo{author}{X.~Cai},
  \bibinfo{author}{X.~Wang}, \bibinfo{author}{H.~Zhang},
  \bibinfo{author}{D.~Wang},
\newblock \bibinfo{title}{Collaborative generative adversarial network with
  visual perception and memory reasoning},
\newblock \bibinfo{journal}{Neurocomputing} \bibinfo{volume}{414}
  (\bibinfo{year}{2020}) \bibinfo{pages}{101--119}.
\bibitem[{Li et~al.(2021{\natexlab{a}})Li, Fan, Wang, Ma, and
  Cui}]{li2021tackling}
\bibinfo{author}{W.~Li}, \bibinfo{author}{L.~Fan}, \bibinfo{author}{Z.~Wang},
  \bibinfo{author}{C.~Ma}, \bibinfo{author}{X.~Cui},
\newblock \bibinfo{title}{Tackling mode collapse in multi-generator {GAN}s with
  orthogonal vectors},
\newblock \bibinfo{journal}{Pattern Recognition} \bibinfo{volume}{110}
  (\bibinfo{year}{2021}{\natexlab{a}}) \bibinfo{pages}{107646}.
\bibitem[{Li et~al.(2021{\natexlab{b}})Li, Xing, Xu, Cai, and
  Cheng}]{li2021attention}
\bibinfo{author}{P.~Li}, \bibinfo{author}{X.~Xing}, \bibinfo{author}{X.~Xu},
  \bibinfo{author}{B.~Cai}, \bibinfo{author}{J.~Cheng},
\newblock \bibinfo{title}{Attention-aware concentrated network for saliency
  prediction},
\newblock \bibinfo{journal}{Neurocomputing} \bibinfo{volume}{429}
  (\bibinfo{year}{2021}{\natexlab{b}}) \bibinfo{pages}{199--214}.
\bibitem[{Zhang et~al.(2019)Zhang, Goodfellow, Metaxas, and
  Odena}]{zhang2019self}
\bibinfo{author}{H.~Zhang}, \bibinfo{author}{I.~Goodfellow},
  \bibinfo{author}{D.~Metaxas}, \bibinfo{author}{A.~Odena},
\newblock \bibinfo{title}{Self-attention generative adversarial networks},
\newblock in: \bibinfo{booktitle}{International Conference on Machine
  Learning}, \bibinfo{organization}{PMLR}, \bibinfo{year}{2019}, pp.
  \bibinfo{pages}{7354--7363}.
\bibitem[{Santurkar et~al.(2018)Santurkar, Tsipras, Ilyas, and
  Madry}]{santurkar2018does}
\bibinfo{author}{S.~Santurkar}, \bibinfo{author}{D.~Tsipras},
  \bibinfo{author}{A.~Ilyas}, \bibinfo{author}{A.~Madry},
\newblock \bibinfo{title}{How does batch normalization help optimization?},
\newblock in: \bibinfo{booktitle}{Advances in Neural Information Processing
  Systems}, \bibinfo{year}{2018}, pp. \bibinfo{pages}{2483--2493}.
\bibitem[{Lian and Liu(2019)}]{lian2019revisit}
\bibinfo{author}{X.~Lian}, \bibinfo{author}{J.~Liu},
\newblock \bibinfo{title}{Revisit batch normalization: New understanding and
  refinement via composition optimization},
\newblock in: \bibinfo{booktitle}{The 22nd International Conference on
  Artificial Intelligence and Statistics}, \bibinfo{year}{2019}, pp.
  \bibinfo{pages}{3254--3263}.
\bibitem[{Schuster and Paliwal(1997)}]{schuster1997bidirectional}
\bibinfo{author}{M.~Schuster}, \bibinfo{author}{K.~K. Paliwal},
\newblock \bibinfo{title}{Bidirectional recurrent neural networks},
\newblock \bibinfo{journal}{IEEE transactions on Signal Processing}
  \bibinfo{volume}{45} (\bibinfo{year}{1997}) \bibinfo{pages}{2673--2681}.
\bibitem[{Kingma and Ba(2015)}]{Kingma2015AdamAM}
\bibinfo{author}{D.~P. Kingma}, \bibinfo{author}{J.~Ba},
\newblock \bibinfo{title}{Adam: A method for stochastic optimization},
\newblock \bibinfo{journal}{CoRR} \bibinfo{volume}{abs/1412.6980}
  (\bibinfo{year}{2015}).
\bibitem[{Heusel et~al.(2017)Heusel, Ramsauer, Unterthiner, Nessler, and
  Hochreiter}]{heusel2017gans}
\bibinfo{author}{M.~Heusel}, \bibinfo{author}{H.~Ramsauer},
  \bibinfo{author}{T.~Unterthiner}, \bibinfo{author}{B.~Nessler},
  \bibinfo{author}{S.~Hochreiter},
\newblock \bibinfo{title}{{GAN}s trained by a two time-scale update rule
  converge to a local {Nash} equilibrium},
\newblock in: \bibinfo{booktitle}{Advances in neural information processing
  systems}, \bibinfo{year}{2017}, pp. \bibinfo{pages}{6626--6637}.
\bibitem[{Paszke et~al.(2019)Paszke, Gross, Massa, Lerer, Bradbury, Chanan,
  Killeen, Lin, Gimelshein, Antiga et~al.}]{paszke2019pytorch}
\bibinfo{author}{A.~Paszke}, \bibinfo{author}{S.~Gross},
  \bibinfo{author}{F.~Massa}, \bibinfo{author}{A.~Lerer},
  \bibinfo{author}{J.~Bradbury}, \bibinfo{author}{G.~Chanan},
  \bibinfo{author}{T.~Killeen}, \bibinfo{author}{Z.~Lin},
  \bibinfo{author}{N.~Gimelshein}, \bibinfo{author}{L.~Antiga}, et~al.,
\newblock \bibinfo{title}{Pytorch: An imperative style, high-performance deep
  learning library},
\newblock in: \bibinfo{booktitle}{Advances in neural information processing
  systems}, \bibinfo{year}{2019}, pp. \bibinfo{pages}{8026--8037}.
\bibitem[{Reed et~al.(2016{\natexlab{a}})Reed, Akata, Yan, Logeswaran, Schiele,
  and Lee}]{10.5555/3045390.3045503}
\bibinfo{author}{S.~Reed}, \bibinfo{author}{Z.~Akata},
  \bibinfo{author}{X.~Yan}, \bibinfo{author}{L.~Logeswaran},
  \bibinfo{author}{B.~Schiele}, \bibinfo{author}{H.~Lee},
\newblock \bibinfo{title}{Generative adversarial text to image synthesis},
\newblock in: \bibinfo{booktitle}{Proceedings of the 33rd International
  Conference on International Conference on Machine Learning - Volume 48},
  ICML'16, \bibinfo{publisher}{JMLR.org}, \bibinfo{year}{2016}{\natexlab{a}},
  p. \bibinfo{pages}{1060–1069}.
\bibitem[{Reed et~al.(2016{\natexlab{b}})Reed, Akata, Mohan, Tenka, Schiele,
  and Lee}]{reed2016learning}
\bibinfo{author}{S.~E. Reed}, \bibinfo{author}{Z.~Akata},
  \bibinfo{author}{S.~Mohan}, \bibinfo{author}{S.~Tenka},
  \bibinfo{author}{B.~Schiele}, \bibinfo{author}{H.~Lee},
\newblock \bibinfo{title}{Learning what and where to draw},
\newblock in: \bibinfo{booktitle}{Advances in neural information processing
  systems}, \bibinfo{year}{2016}{\natexlab{b}}, pp. \bibinfo{pages}{217--225}.
\bibitem[{Szegedy et~al.(2016)Szegedy, Vanhoucke, Ioffe, Shlens, and
  Wojna}]{szegedy2016rethinking}
\bibinfo{author}{C.~Szegedy}, \bibinfo{author}{V.~Vanhoucke},
  \bibinfo{author}{S.~Ioffe}, \bibinfo{author}{J.~Shlens},
  \bibinfo{author}{Z.~Wojna},
\newblock \bibinfo{title}{Rethinking the inception architecture for computer
  vision},
\newblock in: \bibinfo{booktitle}{Proceedings of the IEEE conference on
  computer vision and pattern recognition}, \bibinfo{year}{2016}, pp.
  \bibinfo{pages}{2818--2826}.
\bibitem[{Zhang et~al.(2018)Zhang, Isola, Efros, Shechtman, and
  Wang}]{zhang2018unreasonable}
\bibinfo{author}{R.~Zhang}, \bibinfo{author}{P.~Isola}, \bibinfo{author}{A.~A.
  Efros}, \bibinfo{author}{E.~Shechtman}, \bibinfo{author}{O.~Wang},
\newblock \bibinfo{title}{The unreasonable effectiveness of deep features as a
  perceptual metric},
\newblock in: \bibinfo{booktitle}{Proceedings of the IEEE conference on
  computer vision and pattern recognition}, \bibinfo{year}{2018}, pp.
  \bibinfo{pages}{586--595}.
\bibitem[{Li et~al.(2019)Li, Zhang, Zhang, Huang, He, Lyu, and
  Gao}]{li2019object}
\bibinfo{author}{W.~Li}, \bibinfo{author}{P.~Zhang},
  \bibinfo{author}{L.~Zhang}, \bibinfo{author}{Q.~Huang},
  \bibinfo{author}{X.~He}, \bibinfo{author}{S.~Lyu}, \bibinfo{author}{J.~Gao},
\newblock \bibinfo{title}{Object-driven text-to-image synthesis via adversarial
  training},
\newblock in: \bibinfo{booktitle}{Proceedings of the IEEE Conference on
  Computer Vision and Pattern Recognition}, \bibinfo{year}{2019}, pp.
  \bibinfo{pages}{12174--12182}.

\end{thebibliography}




\end{document}